\pdfoutput=1
\documentclass[11pt]{article}

\usepackage{acl}

\usepackage{times}
\usepackage{latexsym}

\usepackage[T1]{fontenc}
\usepackage[utf8]{inputenc}

\usepackage{microtype}

\usepackage{url}

\usepackage{amsmath}
\usepackage{graphicx}
\usepackage{makecell}
\usepackage{caption}
\usepackage{subcaption}
\usepackage{booktabs}
\usepackage{tabularx}
\usepackage{multirow, makecell}
\usepackage{booktabs}
\usepackage[T1]{fontenc}

\usepackage{color,soul}
\usepackage{enumitem}

\usepackage[linecolor=red,backgroundcolor=yellow!40, textsize=scriptsize]{todonotes}
\usepackage{hyperref}
\usepackage{cleveref}

\crefformat{section}{\S#2#1#3} % see manual of cleveref, section 8.2.1
\crefformat{subsection}{\S#2#1#3}
\crefformat{subsubsection}{\S#2#1#3}

\newcommand{\pgname}{\underline{\textbf{Co}}mmon\underline{\textbf{Se}}nse \underline{\textbf{Co}}ntextualizer}
\newcommand{\pgsname}{{\fontfamily{qag}\selectfont CoSe-Co}}
\newcommand{\dname}{diverse-path search}

\title{\pgsname: Text Conditioned Generative CommonSense Contextualizer}

\author{Rachit Bansal$^1$\thanks{\enskip Work done as an intern at the Media and Data Science Research Lab, Adobe, India} \quad Milan Aggarwal$^2$ \quad Sumit Bhatia$^2$ \\
\textbf{Jivat Neet Kaur$^{3 \ast}$ \quad Balaji Krishnamurthy$^2$} \\
 $^1$~Delhi Technological University \quad 
 $^2$~Media and Data Science Research Lab, Adobe, India \\
 $^3$~Microsoft Research, India \\
\texttt{racbansa@gmail.com, \{milaggar, sumbhati, kbalaji\}@adobe.com} \\
\texttt{jivatneet@gmail.com}}%

\begin{document}
\maketitle
\begin{abstract}
Pre-trained Language Models (PTLMs) have been shown to perform well on natural language tasks. Many prior works have leveraged structured commonsense present in the form of entities linked through labeled relations in Knowledge Graphs (KGs) to assist PTLMs. Retrieval approaches use KG as a separate static module which limits coverage since KGs contain finite knowledge. Generative methods train PTLMs on KG triples to improve the scale at which knowledge can be obtained. However, training on symbolic KG entities limits their applicability in tasks involving natural language text where they ignore overall context. To mitigate this, we propose a \pgname\ (\pgsname) conditioned on sentences as input to make it generically usable in tasks for generating knowledge relevant to the overall context of input text. To train \pgsname, we propose a novel dataset comprising of sentence and commonsense knowledge pairs. The knowledge inferred by \pgsname\ is diverse and contain novel entities not present in the underlying KG. We augment generated knowledge in Multi-Choice QA and Open-ended CommonSense Reasoning tasks leading to improvements over current best methods on CSQA, ARC, QASC and OBQA datasets. We also demonstrate its applicability in improving performance of a baseline model for paraphrase generation task.

% Further, improved performance is seen in low training data regimes which shows \pgsname\ knowledge helps in generalising better.

\end{abstract}

\section{Introduction}
% Following the lines of prior work, we wish to inculcate commonsense reasoning abilities into language models using commonsense knowledge graphs, while retaining maximum knowledge from both worlds. We wish to do so such that this model can take natural language sentences and to contextualise them using commonsense inferences. Further, these inferences are desired to be multi-hop paths consisting of novel unseen entities.

% When faced with natural language

% comfortably\usepackage{}

% Based on experiences and acquired knowledge, humans can employ common sense for better understanding of their environment and improved decision-making.

%_sumit_
% Common sense, as defined by the Merriam-Webster dictionary, is ``\textit{sound and prudent judgment based on a simple perception of the situation or facts.}"  
While dealing with natural language text, commonsense allows humans to expand salient concepts and infer additional information. For example, by reading a sign like \textit{Men at Work} on a road, we implicitly know to slow down our vehicles, look carefully for workers. This implicit process of using common sense to make logical inferences is critical to natural language understanding~\citep{xie2021commonsense}. A natural question to ask then is how we can incorporate common sense in now-ubiquitous language models (LMs)~\citep{devlin-etal-2019-bert, radford2018improving, raffel2019t5}.

There have been various efforts~\citep{bao-etal-2016-constraint, feng-etal-2020-scalable, pgqa} to leverage structured knowledge present in commonsense knowledge
graphs - KGs (we use KG as a shorthand for Commonsense Knowledge Graph) ~\citep{xie2021commonsense}. Such works have primarily focused on either retrieving or generating required knowledge. Retrieval methods rely heavily on structure of downstream task like multi-choice question answering (QA) to leverage knowledge in a KG~\citep{Yasunaga2021QAGNN} and are not applicable beyond a specific task. Further, retrieval can restrict total knowledge that can be garnered since static KGs lack coverage due to sparsity~\citep{bordes2013translating, guu-etal-2015-traversing}. The other body of work addresses this comprising of generative methods that learn commonsense through training a LM on symbolic entities and relations between them in a KG. They have either been designed for KG completion~\citep{comet}, i.e. generate tail entity of a KG triple given head entity and relation, or to generate commonsense paths connecting a pair of entities which suffer from two shortcomings. Firstly, applying such methods in downstream tasks require entity extraction from text as a prerequisite step and secondly, they generate knowledge between entity pairs ignoring overall context of sentence~\citep{pgqa}. Hence, applying such methods is sub-optimal since most NLP tasks comprise of sentences. Further, being trained on entities, applying them directly on sentences is infeasible and lead to train-inference input type mismatch.

To address these limitations, we propose \pgname\ - \textbf{\pgsname}, 
a generative framework which generates relevant commonsense knowledge given natural language sentence as input. 
% Being conditioned on sentences 
% makes \pgsname\ generically applicable by
% enabling it to dynamically select entities/phrases from an input sentence 
% most relevant to the task, and further 
% expand them through novel commonsense knowledge.
We condition it on sentences to make it learn to incorporate overall text context and enable it to dynamically select entities/phrases from an input sentence 
as well as output novel yet relevant entities 
% (which are not present in KG) 
as part of commonsense inferences generated.
%
% Figure \ref{fig:figure_1} shows how \pgsname\ could be used to generate commonsense inferences for a variety of tasks.
%
% Figure \ref{fig:figure_1} shows few commonsense inferences generated by \pgsname\ for three tasks. 
% Relation vocabulary is determined based on schema of KG used for training \pgsname. 
We consider commonsense knowledge in the form of paths, i.e., sequence of entities connected through relations.
% To achieve this, 
We first create sentence-path paired dataset 
% (since no such data exists) 
by - 1) sampling paths from an underlying KG; 2) sampling a subset of entities from a path; and 3) retrieving \& filtering sentences 
(from a sentence corpus) that
% contain sampled entities and
are semantically similar to the path. The paired data is then used to train a generative language model to generate a path given a sentence as input.

To analyse the usefulness of generated commonsense, we augment it in various downstream tasks. The reasoning ability of NLP systems is commonly analysed using QA. Hence, we choose two such tasks: 1) Multi-Choice QA, where given a question and set of choices, the model has to identify the most appropriate answer choice. 
However, often more than one choice is a suitable answer. To mitigate this, 2) OpenCSR (Open-ended CommonSense Reasoning)~\citep{lin-etal-2021-differentiable} was proposed, where each question is labeled with a set of answers which have to be generated without choices. We also show applicability of \pgsname\ in improving performance on paraphrase generation task (\cref{sec:other_tasks}). 

Our contributions can be summarised as:

% propose sentence conditioned generative task agnostic (sec 3.) -- 1, no dataset, devise methodology (s 3.1) and release -- 2 , novel entities, relevant diverse, kg construction better (4.2) -- 3, finally results (% improvements) -- 4.3-4.5

%  We also provide a method to create paired sentence-path data for training \pgsname.

% \item Our framework adapts to text input in downstream tasks since it does not suffer from train-inference input distribution shift which other generative commonsense LMs face.

%  since it is not strictly tied to the entities in the KG used for training it owing to its generative nature

% since no dataset comprising of such mappings exists
% \comment{
\begin{enumerate}[noitemsep,nolistsep]
    \item We propose a \pgname\ (\pgsname) to generate knowledge relevant to overall context of given natural language text. \pgsname\ is conditioned on sentence as input to make it generically usable in tasks without relying on entity extraction.
    % \item We devise a method to create novel sentence - relevant commonsense pairs to train \pgsname\ (\cref{sec:methodology}) as no such dataset exists. We release dataset and trained \pgsname\ model \href{https://drive.google.com/drive/folders/19kGz-fGkYM-FAtcT47QiaUclpNd7ZAXC?usp=sharing}{here}.
    \item We devise a method to extract sentence-relevant commonsense knowledge paths and create the first sentence-path paired dataset.
    % \pgsname\ (\cref{sec:methodology}) as no such dataset exists. 
    We release the dataset and make it available to the community along with the trained models and corresponding code\footnote{\url{https://linktr.ee/coseco}}.
    % \footnote{\url{https://drive.google.com/drive/folders/19kGz-fGkYM-FAtcT47QiaUclpNd7ZAXC?usp=sharing}}
    % The code will be released at this link.\footnote{\url{https://github.com/nl-reasoning/CoSe-Co}}
    % \href{https://drive.google.com/drive/folders/19kGz-fGkYM-FAtcT47QiaUclpNd7ZAXC?usp=sharing}{here}.
    % \item Since \pgsname\ is based on a generative LM, it infers relevant and diverse paths which contain novel entities not present in the underlying KG (\cref{sec:analyse_p}). % Further, we show that it can be used for the task of KG completion performing better than previous method.
    \item Since \pgsname\ is based on generative LM, it infers relevant and diverse knowledge containing novel entities not present in the underlying KG (\cref{sec:analyse_p}).
    Augmenting generated knowledge in Multi-Choice QA (\cref{sec:down_csqa}) and OpenCSR (\cref{sec:down_opencsr}) tasks leads to improvements over current SoTA methods. Further, it is observed that \pgsname\ helps in generalising better in low training data regime. % Further, using \pgsname\ paths results in better performance in low training data regimes indicating that they help in generalizing better. 
    % Especially 
    % Significant improvements in low-data regime indicates the effect of \pgsname\ on generalization.
\end{enumerate}
% }
% Further, we demonstrate the generisability of our framework by proposing a mechanism to perform zero-shot QA for OpenCSR where it performs comparable to supervised benchmarks.

% task agnostic in contributions summary
% leveraging the power of language model which are usually pre-trained on natural language sentences to combine LM's generic semantic understanding with common sense KG
% generic in the sense it is not restricted to KG (criticise QA-GNN more in related work)
% not only linking specific entities

% This is contrary to the properties of commonsense knowledge, which in itself is both expressive and generic. Beyond question-answering \cite{sap-etal-2019-social, talmor-etal-2019-commonsenseqa, obqa}, which is the primary task through which progress of LM + common sense KGs systems is currently measured, commonsense has an obvious role to play in a vast variety natural language processing tasks, like machine translation and dialogue systems \cite{dimesions_of_commonsense}.

% % need to think of a better name for the approach -  mainly path generator

% In order to preserve these attributes of commonsense knowledge in language tasks, we propose to make inferences from natural text such that selective extraction from either of the two is not required. Specifically, we design \pgsname : \pgname, that can take natural sentences in their original form, and make inferences about the contained concepts with respect to the larger sentence context.   \\
% \\
% \rb{Summarise contributions point-wise}
% \vspace{-2mm}
\section{Related Work}
\label{sec:rel_work}

Commonsense Knowledge Graphs (KGs) are structured knowledge sources comprising of entity nodes in the form of symbolic natural language phrases connected through relations~\citep{conceptnet, atomic, ilievski2021cskg, zhang2020aser}. The knowledge in KGs is leveraged to provide additional context in NLP tasks~\citep{bao-etal-2016-constraint, sun-etal-2018-open, lin-etal-2019-kagnet} and perform explainable structured reasoning~\citep{Ren*2020Query2box:, NEURIPS2020_e43739bb}. Additionally, a variety of Natural Language Inference (NLI) and generation tasks requiring commonsense reasoning have been proposed over the years~\citep{zellers-etal-2018-swag, talmor-etal-2019-commonsenseqa, sap-etal-2019-social, lin-etal-2020-commongen, lin-etal-2021-differentiable, lin2021riddlesense}. Pre-trained language models (PTLMs)~\citep{devlin-etal-2019-bert} trained over large text corpus have been shown to posses textual knowledge~\citep{whatLMKnow, lmaskb, knowledgePackLM} and semantic understanding~\citep{li-etal-2021-implicit}. Consequently, they have been used for reasoning where they perform well to some extent~\citep{bhagavatula2019abductive, huang-etal-2019-cosmos}. However, it remains unclear whether this performance can be genuinely attributed to reasoning capability or if it is due to unknown data correlation~\citep{mitra2019exploring, niven-kao-2019-probing, kassner-schutze-2020-negated, zhou2020rica}.

\setlength{\belowcaptionskip}{-12pt}
\begin{figure*}[h]
    \centering
    \includegraphics[width=0.95\textwidth,height=1.65in]{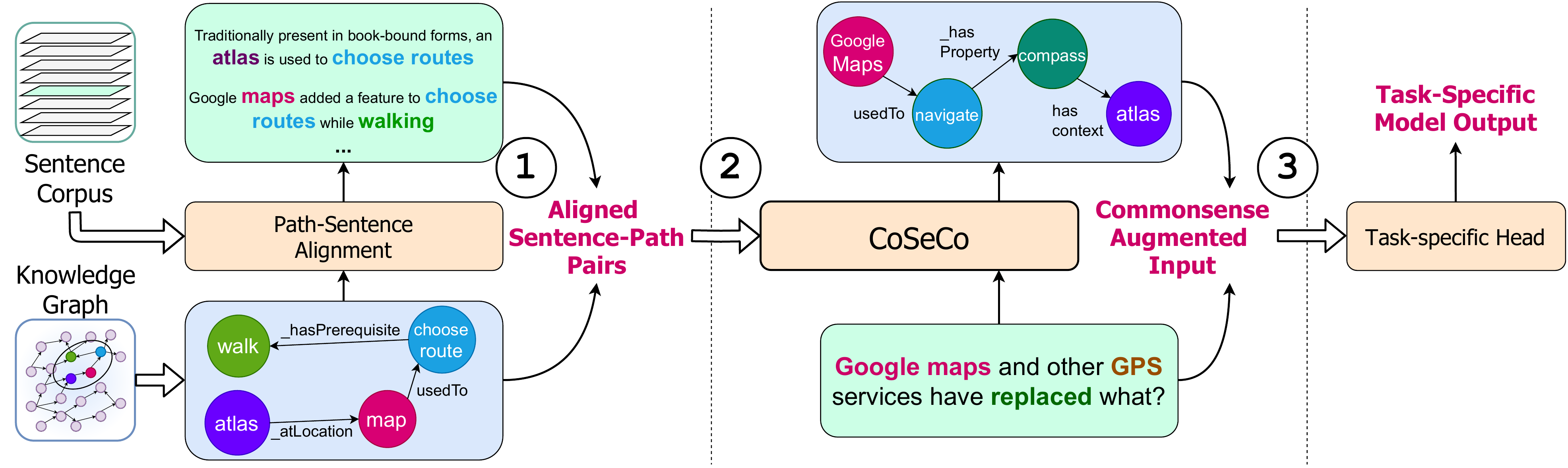}
    \caption{Our proposed approach consists of: (1) \textbf{Path to Sentence Alignment} to create the training data for \pgsname, (2) Training a \textbf{\pgname}\ (\pgsname) to generate commonsense inferences relevant to a given natural language sentence. \pgsname\ can be used to infer knowledge in downstream task.}
    \label{fig:approach}
\end{figure*}

% \vspace{-0.5mm}
Due to this, various LM + KG systems have been explored~\citep{feng-etal-2020-scalable, wang2019improving, Lv_Guo_Xu_Tang_Duan_Gong_Shou_Jiang_Cao_Hu_2020} to combine broad textual coverage of LMs with KG's structured reasoning capability. Early works on KG guided QA retrieve sub-graph relevant to question entities but suffer noise due to irrelevant nodes~\citep{bao-etal-2016-constraint, sun-etal-2018-open}. Hybrid graph network based methods generate missing edges in the retrieved sub-graph while filtering out irrelevant edges~\citep{yan2020learning}. Graph Neural Networks (GNNs) have been used to model embeddings of KG nodes~\citep{wang2020entity}. More recently, \citet{Yasunaga2021QAGNN} proposed an improved framework (QA-GNN) leveraging a static KG by unifying GNN based KG entity embeddings with LM based QA representations. Although, such frameworks extract relevant evidence from a KG, it undesirably restricts knowledge that can be garnered since knowledge source is static and might lack coverage due to sparsity~\citep{bordes2013translating, guu-etal-2015-traversing}. Contrarily, we train a generative model on a given KG to enable it to dynamically generate relevant commonsense inferences making it more generalizable and scalable.

% \vspace{-1mm}
\citet{comet} cast commonsense acquisition by LMs as KG completion. They propose COMET, a GPT~\citep{gpt} based framework to generate tail entity given head and relation in a KG triple as input. Owing to training on symbolic KG nodes, using COMET in downstream tasks involving natural language text is not straightforward. Specifically, it requires extracting entities from text as a prerequisite~\citep{becker-EtAl:2021:IWCS}. Further, training on single triples makes its application in tasks requiring multi-hop reasoning challenging due to large relation search space~\citep{bosselut2020dynamic}. To address this, Path Generator (PGQA) was proposed to generate commonsense paths between entities pair~\citep{pgqa}. Designed for multi-choice QA, they extract question entities and generate paths between each question entity and answer choice pair. Even though generated paths are multi-hop, training on entities limits applying it directly on sentences due to train-inference input type mismatch. Further, being conditioned only on question-choice entity pairs, paths are generated ignoring overall question context. To mitigate this, we design \pgsname\ as a generic framework to dynamically generate multi-hop commonsense inference given natural language sentence as input. Separately, retrieval methods have been explored to search relevant sentences to generate text corresponding to concepts~\citep{wang-etal-2021-retrieval-enhanced}. Different from this task, we retrieve sentences relevant to paths in a KG to create paired sentence-path data.
% \vspace{-2mm}
% \begin{figure*}[t]
%     \centering
%     \includegraphics[width=\textwidth]{images/figure_2.pdf}
%     \caption{Our proposed approach consists of three main steps: (1) \textbf{Path to Sentence Alignment} to create the training data for \pgsname, (2) Training a \textbf{\pgname}\ (\pgsname) to generate commonsense inferences relevant to a given natural language sentence, and (3) \textbf{Augmenting inferred knowledge} to the input as additional context in a downstream task.}
%     \label{fig:approach}
% \end{figure*}
% \vspace{-2mm}
\setlength{\belowcaptionskip}{-12pt}
\begin{figure*}[t]
    \centering
    \includegraphics[width=0.95\textwidth]{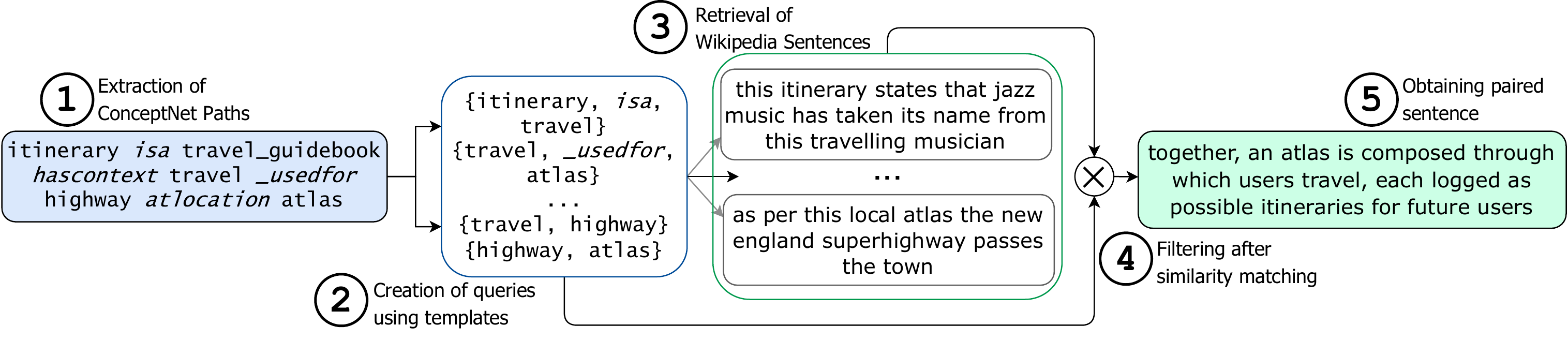}
    \caption{Obtaining the sentence-path paired dataset. We begin with paths from the knowledge graph and employ a two-step matching and filtering process to obtain relevant paired sentences from the given text corpora. Here we accompany each step with corresponding examples that we observed.}
    \label{fig:s_to_p_ill}
\end{figure*}

% \vspace{-1mm}
\section{Proposed \pgsname\ Framework} \label{sec:methodology}
% \vspace{-1mm}
\paragraph{Problem Setting}
Given a commonsense knowledge graph $\mathcal{G} = (\mathcal{E}, \mathcal{R})$, where $\mathcal{E}$ is the set of entity nodes and $\mathcal{R}$ is the set of labeled directed relational edges between entities, we aim to model a \pgname\ (\textbf{\pgsname}) which generates a set of commonsense inferences in the form of paths derived using $\mathcal{G}$, that are relevant to a natural language text given as input. It is desirable that such a generative commonsense knowledge model should be generic, task agnostic, and takes into account the overall context of language input while generating commonsense. Since most tasks comprise of text in the form of sentences, we model the input to \pgsname\ as a sentence. In order to train such a model, a dataset is required which comprises of mappings of the form $\{(s_1, p_1), (s_2, p_2), ..., (s_N, p_N)\}$, where $s_j$ and $p_j$ are relevant sentence-commonsense inference path pair. 
However, no existing dataset consists of such mappings. 
To bridge this gap, we first devise a methodology to create a dataset $\mathcal{D}$ comprising of sentences paired with relevant commonsense inference paths. 
Broadly, we first extract a large corpus $\mathcal{C}$ constituting sentences $\{s_1, s_2, ..., s_{|C|}\}$. Subsequently, we sample a set of paths $\mathcal{P} = \{p_1, p_2, ..., p_{|\mathcal{P}|}\}$ from $\mathcal{G}$ such that each $p \in \mathcal{P}$ is of the form $p = \{e_1, r_1, e_2, r_2, ..., e_{|p|+1}\}$, where $e_i \in \mathcal{E}$ and $r_i \in \mathcal{R}$. For each $p \in \mathcal{P}$, a set of contextually and semantically relevant sentences $S \subset \mathcal{C}$ is retrieved and mapped to $p$. We then train a generative LM based commonsense knowledge model using $\mathcal{D}$. During inference, given a sentence $s'$, it generates commonsense paths of the form $p' = \{e'_1, r'_1, e'_2, r'_2, ..., e'_{|p'|+1}\}$ such that $e'_i \in \mathcal{E'}$ and $r'_i \in \mathcal{R}$. Here, $\mathcal{E'} = \mathcal{E} \cup \mathcal{E}_{novel}$ where $\mathcal{E}_{novel}$ are novel entities not present in $\mathcal{G}$. These include phrases present in an input sentence but not in $\mathcal{E}$ as well as entirely novel entities which the pre-trained LM based backbone enables it to generate through transfer learning. The generated commonsense inference paths from \pgsname\ can then be used to augment context in downstream tasks. An overview of our framework is shown in Figure \ref{fig:approach}. 
% We now explain each step in detail.

% \vspace{-1mm}
\subsection{Sentence-Path Paired Dataset Creation}
\label{sec:s_to_p}
% \vspace{-1mm}
In order to train \pgsname, we create a novel dataset comprising of related sentence-commonsense path pairs. To obtain set $\mathcal{P}$, we perform random walk in $\mathcal{G}$ to extract multi-hop paths of the form $p = \{e_1, r_1, e_2, r_2, ..., e_{|p|+1}\}$, where the number of hops, denoted as path length $|p|$, is in range $[l_1, l_2]$. To avoid noisy paths which do not convey useful information, 
% we employ heuristics to filter generic relations 
we employ relational heuristics 
% in $\mathcal{R}$~\citep{pgqa}. 
in $\mathcal{P}$ (described in \cref{ap:rel_heur}).
Separately, the sentence corpus $\mathcal{C}$ is indexed using \href{https://solr.apache.org/}{Apache Solr} which is queried to retrieve sentences relevant to a path. We now explain this in detail.
\indent Broadly, we map each path $p \in \mathcal{P}$ to a set of sentences $S \subset \mathcal{C}$ based on semantic similarity and overlap between entities in $p$ and sentences. For this, consider a path $p = \{e_1, r_1, e_2, ..., e_{|p|+1}\}$. To ensure that retrieved sentences are similar to $p$, we devise two types of query templates - $Q1$ and $Q2$ which are used to create multiple queries per path while querying Solr. We design $Q1$ to capture relation information between entities in $p$ in addition to entities themselves. Specifically, we extract non-contiguous entity-relation triples of the form $\{(e_i, r_i, e_{i+2})\}$ and $\{(e_i, r_{i+1}, e_{i+2})\}$. 
% One thing to note is that
Here, we do not query entire path while retrieving sentences to ensure better coverage since we observed that \textbf{no sentence exists} which contains \textbf{all entities and relations} present in a given path. In $Q2$, we extract queries comprising of connected entities pairs $\{(e_i, e_{i+1})\}$. For each query $q$ obtained from $p$ according to $Q1$ and $Q2$, we query Solr and select sentences containing \textbf{entities} 
%(relation as well for queries of type $Q1$) 
present in $q$. Subsequently, we rank retrieved sentences based on similarity between sentence embedding and embedded representation of the corresponding query $q$ (including the relation in case of $Q1$). The embeddings are obtained using SBERT~\citep{sbert} since it is trained on siamese objective to learn semantically meaningful representations. 
Based on the ranking, we retain a maximum of top K’ (= 10) sentences to ensure most semantically relevant sentences-path pairs are obtained and also to prevent \pgsname\ from getting biased towards generating particular paths. One thing to notice is that even though sentences are retrieved using templated sub-parts within a path, the retrieved sentences are finally paired up with the \textbf{entire path} and later used to train a generative commonsense model that learns to generate the path given sentence as input.
Figure \ref{fig:s_to_p_ill} illustrates the entire sentence-path pairing process using an example from the dataset.

% \vspace{-1mm}
Using queries of type $Q1$ templates enables us to retrieve sentences that are relatively more semantically related to the overall path. For instance, consider a path `violin  \textit{hasproperty}  strings  \textit{\_hasprequistite}  guitar  \textit{atlocation}  concert'. Sentences retrieved using queries like $\{$strings, \textit{atlocation}, concert$\}$ (of the form $(e_i, r_{i+1}, e_{i+2})$) are more likely to be related to other entities in the path such as `guitar'. Further, sentences that contain entities that are not directly connected in the corresponding path induce an inductive bias in \pgsname\ to generate paths that consist of intermediate entities which connect them.
We perform ablations regarding query templates in \cref{sec:abl}. We study quality of the generated dataset to check for possible data leaks and relevance between sentence-path pairs

% as follows:

% \begin{itemize}
We determine the extent of n-gram overlap between questions in the CSQA test set and sentences in our sentence-path training set as indicators of any possible data leakage. For this, we obtain the set of n-grams in a question, determine the sentence in the training set with which the question has maximum matching n-grams and divide the matching n-gram count with the total number of n-grams in the question. Finally, this fraction is averaged over all the questions in the test split of CSQA. Following this scheme, an overlap of 0.15 is observed for 1-grams, 0.07 for 2-grams, 0.002 for 3-grams, and 0.00 for 4-grams which shows that the extent of overlap is very less (on a scale of 0 to 1). Further, we noted that 1-gram overlap does not necessarily indicate leakage. For instance, consider CSQA test question - ‘If a person is tired how can they be refreshed?’. Even though, it has matching 1-grams with the sentence- ‘a person may feel tired without having engaged in any physical activity’, but it can be noted that they have an entirely different context. From the low n-gram overlap values, we conclude that extent of leakage is negligible.

To gauge the degree of relevance between the final set of sentence-path pairs, we measure the cosine similarity between the S-BERT embeddings of the complete path and the corresponding sentence in the dataset. We observe a high normalized cosine similarity score of 0.783 when averaged over all sentence-path pairs in training dataset which shows that sentence and corresponding path pairs are semantically related.
% \end{itemize}

% Please refer to Appendix \ref{app:disc_data} for further discussion on quality of dataset obtained.

% Our query template based sentence retrieval is aimed at enabling \pgsname\ to learn to extrapolate and interpolate entities while generating path given a sentence as input. 
% \vspace{-2mm}
\subsection{Sentence $\rightarrow$ Commonsense Generator}
\label{sec:s_to_p_pg}
The sentence-commonsense paired dataset $\mathcal{D}$ obtained in \cref{sec:s_to_p} is used to train a path generator model \pgsname$_\theta$\ to generate commonsense inference path $p$ relevant to the input sentence $s$. For this, we initialise the parameters $\theta$ of \pgsname\ with weights of a generative pre-trained LM as backbone (eg. T5, GPT etc). Consider T5-base \citep{raffel2019t5} as backbone, given a sentence $s = \{x^{s}_1, x^{s}_2, ..., x^{s}_{|s|}\}$ comprising of a sequence of tokens, it is processed by T5 encoder $E_{\theta_1}$ to give a sequence of outputs $O_E = \{o^{E}_{1}, o^{E}_{2}, ..., o^{E}_{|s|}\}$. T5 decoder $D_{\theta_2}$ is trained to sequentially generate the corresponding path tokens $p = \{x^{p}_1, x^{p}_2, ..., x^{p}_{N}\}$. During the decoding phase at time step $t$, $D_{\theta_2}$ is jointly conditioned on encoder outputs $O_E$ and past tokens $x^{p}_{<t}$ in the path $p$ while generating current path token $x^{p}_t$. $E_{\theta_1}$ and $D_{\theta_2}$, where $\theta = \theta_1 \bigcup \theta_2$, are jointly optimized by minimizing loss $\mathcal{L}$:
% \vspace{-1mm}
\begin{center}
    $\mathcal{L} = -\sum_{t=1}^{N} \log P(x^{p}_t | x^{p}_{<t}, O_E) $, where $P(x^{p}_t | x^{p}_{<t}, O_E) =$ \pgsname$_\theta(s, x^{p}_{<t})$
\end{center}
% \vspace{-1mm}

\noindent We design a variant where given a sentence-path pair, we randomly select an entity that co-occurs in sentence and path and mask it in the sentence. 
% A hyper-parameter $p_{mask}$ depicting the probability of whether the sentence is to be masked is used otherwise original sentence is provided as input. %
Whether a sentence is masked during training is controlled by a probability $p_{mask}$.
The model is then trained to generate path containing masked entity given masked sentence as input. The intuition is to enforce \pgsname\ to capture context better through
% understand the sentence context better, 
identifying masked entity during path generation. We discuss and perform ablations to compare masked \pgsname\ with varying values of $p_{mask}$ in \cref{sec:abl}. Separately, we discuss and observe that using GPT-2 as backbone LM for \pgsname\ performs similar to T5-base in Appendix \ref{app:gpt_experiments}.
% \vspace{-1mm}

% \vspace{-1.5mm}
\subsection{Path Decoding During Inference}
% \vspace{-0.5mm}
As in most sequence generation tasks, teacher forcing is used to train the model, while a decoding strategy is used to generate diverse outputs during inference~\citep{NIPS2017_3f5ee243}. To maximise contextual knowledge obtained from paths for each sentence in a downstream task, we generate multiple paths. To improve diversity between paths while not losing relevance, we implement a path-specific variant of beam search, \textit{\dname}. Diversity is ensured by sampling top-$k$ most probable tokens at first generation step followed by decoding most probable sequence for each of them, thus returning $k$ paths. This approach is motivated by observation that when generating a path, initial entity guide overall decoding of path.
% \vspace{-3mm}
% \vspace{-0.5mm}

\section{Experiments and Evaluation} \label{sec:down}
% \vspace{-1mm}
% For most of our experiments, 
\subsection{Implementation Details}
\label{sec:imp_details}
% \vspace{-1mm}
% In this section, we discuss some implementation details related to \pgsname training.

\setlength{\belowcaptionskip}{-4pt}
\begin{figure*}[t]
    \begin{subfigure}{0.475\textwidth}
        \centering
        \includegraphics[width=0.80\textwidth,height=1.25in]{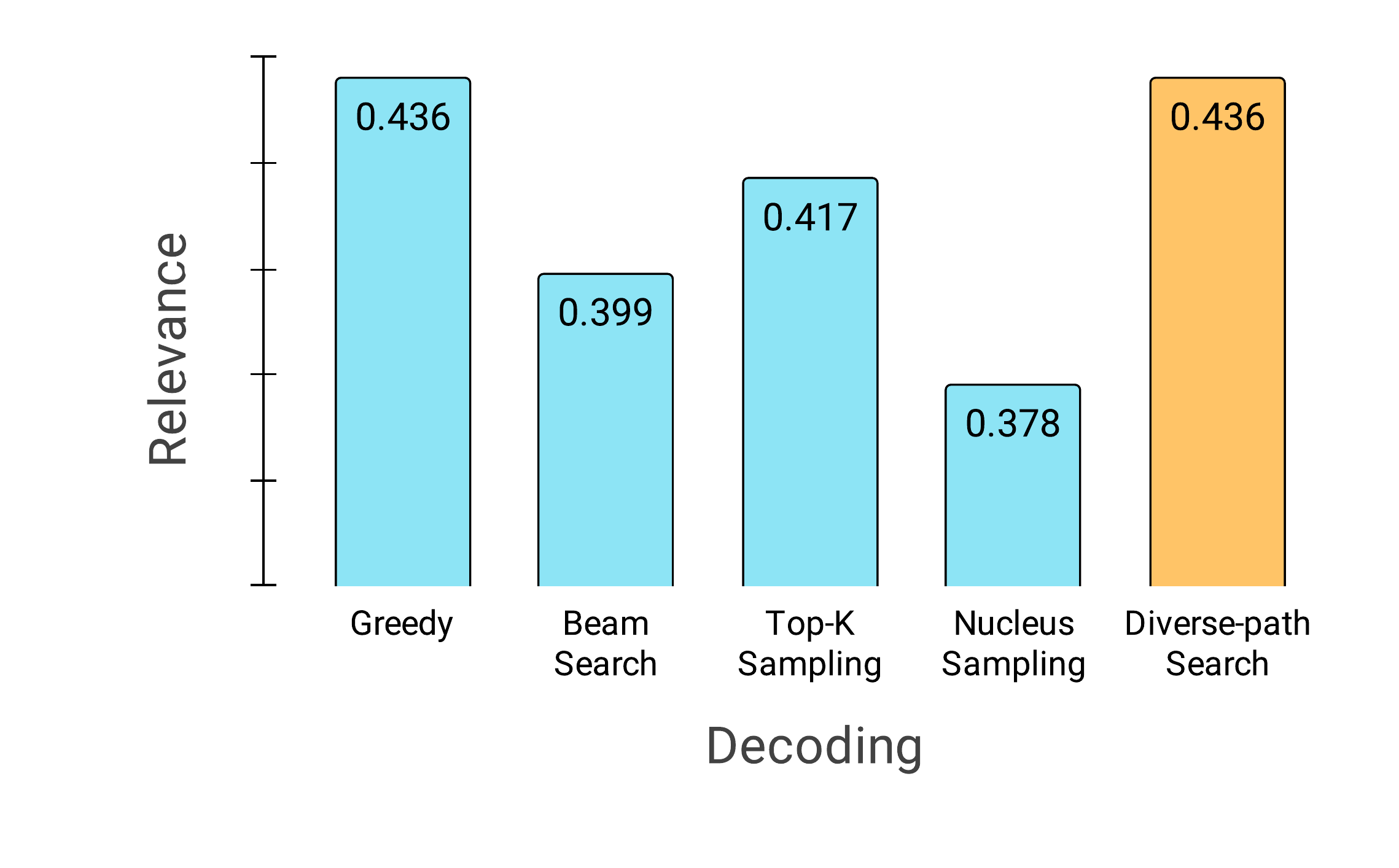}
        % \vspace{-0.5cm}
        \caption{\textbf{Relevance}: BLEU score of generated paths computed using ground truth paths}
        \label{fig:pg_relevance}
    \end{subfigure}
    \hspace{0.03\textwidth}
    \begin{subfigure}{0.475\textwidth}
        \centering
        \includegraphics[width=0.80\textwidth,height=1.25in]{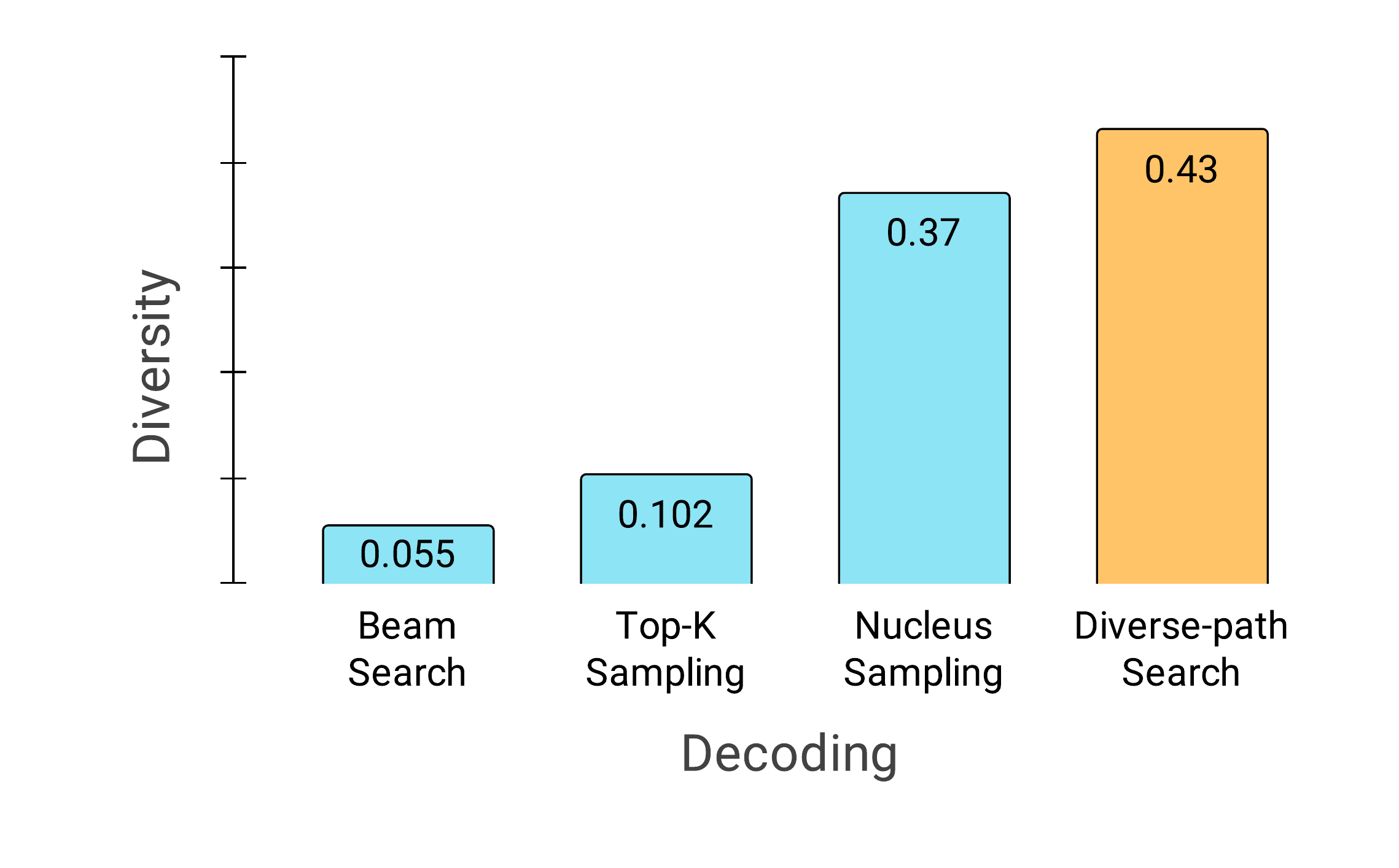}
        % \vspace{-0.5cm}
        \caption{\textbf{Diversity}: Compliment of fraction overlap between top-$5$ sampled paths.}
        \label{fig:pg_diversity}
    \end{subfigure}
    \caption{Analysis and comparison of generated paths across different decoding strategies}
    \label{fig:pg_eval}
\end{figure*}

% In order to evaluate the generalizability, we consider two considerably different question-answering tasks of multi-choice commonsense question-answering and open-ended commonsense reasoning. For the former we train and evaluate the overhead framework on the Commonsense Question-Answering (CSQA;~\citep{talmor-etal-2019-commonsenseqa}) benchmark, while for the latter we evaluate on the recently released Open-Ended Commonsense Reasoning (OpenCSR;~\citep{lin-etal-2021-differentiable}) benchmark. 

% \rb{Should maybe have a figure encompassing our frameworks for both the tasks (at least for CSQA).}
% \colorbox{yellow}{
% \textcolor{ProcessBlue}{\textbf{

\setlength{\belowcaptionskip}{-12pt}
\begin{table*}[t]
\centering
\resizebox{0.98\textwidth}{!}{%
\begin{tabular}{@{}ll@{}}
\toprule
\textbf{Input} &\textbf{\pgsname\ Outputs} \\ \midrule
% \multirow{3}{*}{\begin{tabular}[c]{@{}l@{}}What do people aim to do \\at work?\end{tabular}} & (work\_at\_home \textit{\_capableof} people \textit{desires} work \textit{\_hasprerequisite} \textcolor{cyan}{\textbf{earning\_living}} \textit{causes} \colorbox{yellow}{get\_paid}\_by\_job) \\
%  & (get\_money\_from \textit{\_capableof} people \textit{desires} work \textit{\_hassubevent} \colorbox{yellow}{enjoying\_day}) \\
%  & (have\_to\_work \textit{\_capableof} people \textit{desires} work \textit{causes} getting\_job \textit{hasprerequisite} payrolls\_and\_\textcolor{cyan}{\textbf{paying\_bills}}) \\
%  \midrule
\multirow{3}{*}{\begin{tabular}[c]{@{}l@{}}What do people typically do \\while playing guitar?\end{tabular}} & (playing\_guitar \textit{causes} \colorbox{yellow}{singing} \textit{usedfor} people \textit{capableof} feeling\_sad) \\
 & (playing\_guitar \textit{hassubevent} \colorbox{yellow}{sing} \textit{\_causesdesire} singing \textit{\_occupation} musician \textit{genre} \textcolor{cyan}{\textbf{folk\_rock}}) \\
 & (play\_guitar \textit{\_usedfor} guitar \textit{atlocation} \textcolor{cyan}{\textbf{symphony\_halls}}\_or\_musical\_instruments\_or\_bands \textit{\_atlocation} people ) \\
 \midrule
 \multirow{3}{*}{\begin{tabular}[c]{@{}l@{}}Where are you likely to \\find a hamburger?\end{tabular}} & (burger \textit{\_isa} hamburger \textit{atlocation} \colorbox{yellow}{fast\_food\_restaurant} \textit{usedfor} eating\_food) \\
 & (\colorbox{yellow}{burger\_king} \textit{\_usedfor} hamburger \textit{atlocation} fast\_food\_restaurant \textit{isa} place \textit{capableof} \textcolor{cyan}{\textbf{take\_car\_for\_drive}}) \\
 & (fast\_food\_restaurant \textit{\_isa} \colorbox{yellow}{taco\_bell} \textit{product} hamburger \textit{madeof} \textcolor{cyan}{\textbf{wheat\_flour\_and\_salt}}) \\
 \midrule
 \multirow{3}{*}{\begin{tabular}[c]{@{}l@{}} In what Spanish speaking \\North American country can \\you get a great cup of coffee?\end{tabular}} & (\colorbox{yellow}{bretagne} \textit{partof} north\_america \textit{\_atlocation} cup\_of\_coffee \textit{hascontext} usa \textit{isa} country) \\
 & (\textcolor{cyan}{\textbf{hot\_beverage}} \textit{\_isa} coffee \textit{atlocation} cup\_of\_coffee \textit{hascontext} north\_america \textit{\_partof} \colorbox{yellow}{grenada}) \\
 & (good\_coffee \textit{hasa} \textcolor{cyan}{\textbf{caffiene\_in\_milk\_and\_sugar}} \textit{atlocation} in\_\colorbox{yellow}{canada}) \\
\bottomrule
\end{tabular}%
}
\caption{Commonsense paths generated by \pgsname\ for questions in \textit{CommonsenseQA} data. Potential answers observed in path itself are \colorbox{yellow}{highlighted}, context-enriching concepts are \textcolor{cyan}{\textbf{coloured}}.}
\label{table:path_mask_eval_1}
\end{table*}

We choose Wikipedia as the sentence corpus $\mathcal{C}$,
and ConceptNet~\citep{conceptnet} as the knowledge graph $\mathcal{G}$.
% \pgsname\ is built upon T5-base~\citep{raffel2019t5}, which is a transformer~\citep{NIPS2017_3f5ee243} based generative LM pre-trained on Colossal Clean Crawled Corpus. \pgsname\ consists of $12$ encoder-layers, each with $12$ attention heads, a hidden-state of size $768$, and feed-forward size of $3072$ - resulting in $220$M parameters. 
The subset of \href{https://archive.org/download/enwiki-20181220}{Wikipedia} that we use comprises of $\sim$5M articles, from which we extract $\sim$92.6M sentences. 
ConceptNet comprises of $\sim$8 million nodes as concepts linked through 34 unique commonsense relations with $\sim$21 million links in total. 
We sample $\sim$28M paths that have a length $|p|$ in the range $l_1=2$ and $l_2=5$. We obtain a total of $\sim$290K sentence-path pairs. \pgsname\ is trained until validation loss across an epoch does not increase, with maximum number of epochs $=5$. $p_{mask}$ is set to $0.33$ based on tuning on CSQA dev set and number of paths per sentence $k=5$ during inference. AdamW optimizer~\citep{adamw2017} is used to train parameters with a learning rate of $5e-4$, weight decay of $0.01$ and epsilon of $1e-8$ using a single A-100 GPU with batch-size $8$ and $4$ gradient accumulation steps. 

% \vspace{-1mm}
\subsection{Analysing Generated Paths}
\label{sec:analyse_p}
% \vspace{-1mm}
% The paths generated using \pgsname\ are compared with the ground truth paths on an unseen split of the dataset. The evaluation is done using BLEU score on triples, i.e., each triple in the path is considered as one uni-gram token.

% Before demonstrating the effectiveness of \pgsname\ on downstream tasks, 

We analyse quality of generated paths on three aspects - \textit{Relevance}, \textit{Diversity} and \textit{Novelty}, evaluated on test split of our sentence-path dataset. We estimate \textbf{Relevance} by treating each triple in generated and ground truth paths (for a given test sample) as one uni-gram followed by determining BLEU score~\citep{papineni2002bleu} between them. To estimate \textbf{Diversity}, we extract top-$k=5$ paths for each sentence, consider each pair combination amongst them and estimate fractional overlap (intersection over union of set of path entities) between them. Compliment of overlap ($1 - overlap$) followed by mean over entire test split conveys how diverse paths are. Figure \ref{fig:pg_eval} shows corresponding results. It is observed that paths generated using nucleus sampling are diverse but lack relevance, while an opposite trend is observed for top-k sampling. \textit{Diverse-path search} provides best balance between relevance ($0.436$) and diversity ($0.43$). We estimate \textbf{Novelty} as a fraction of total entities in a generated path that are not present in any training path followed by averaging over test split. \pgsname\ attains a novelty of $23.28\%$ which shows that good fraction of entities in generated path are novel. Further discussion on the quantitative analysis of generated paths can be found in \cref{app:disc_gen_paths}. Table \ref{table:path_mask_eval_1} shows a few examples of generated paths. \pgsname\ generates paths contextually relevant to question in addition to inferring novel entities. 
% Further, we perform KG completion (predicting tail entity given head entity and relation of a KG triple) using \pgsname\ since it generates paths which essentially comprise of triples. We compare the performance with COMET~\citep{comet}. We consider test split of sentence-path dataset comprising of $11,264$ paths and extract triples. We filter out triples appearing in training paths of \pgsname\ and train set triples of COMET yielding $717$ test triples in total. \pgsname\ achieves an accuracy of $24.12\%$ which is significantly better than COMET which provides accuracy of $9.76\%$.

% \vspace{-2mm}
\subsection{Multi-Choice Question Answering} \label{sec:down_csqa}
%%% CSQA InHouse Split Results
\setlength{\belowcaptionskip}{-4pt}
\begin{table*}[htbp]
\centering
\resizebox{1.0\textwidth}{!}{
\begin{tabular}{lcccc}
    \toprule
    \multicolumn{1}{c}{} & \multicolumn{1}{c}{20\% Train} & \multicolumn{1}{c}{60\% Train} & \multicolumn{2}{c}{100\% Train} \\
    \multicolumn{1}{c}{\textbf{Methods}} & \textbf{IHtest} (\%) & \textbf{IHtest} (\%) & \textbf{IHdev} (\%) & \textbf{IHtest} (\%) \\
    \midrule  
    {T5-base (w/o KG) \scalebox{0.7}{\citep{raffel2019t5}}}  & --  & -- & 61.88~($\pm$0.08) & 57.34~($\pm$0.21) \\
    % {T5-large (w/o KG) \scalebox{0.7}{\citep{raffel2019t5}}} & -- & -- &  69.81~($\pm$1.02) & 67.80~($\pm$0.83) \\
    \midrule  
    RoBERTa-large (w/o KG)  & 46.25~($\pm$0.63)  & 52.30~($\pm$0.16) & 73.07~($\pm$0.45) & 68.69 ($\pm$0.56) \\
    \midrule
    + {RGCN \scalebox{0.7}{\citep{schlichtkrull2018modeling}}} & 45.12~($\pm$0.69)  & 54.71~($\pm$0.37) & 72.69~($\pm$0.19) & 68.41~($\pm$0.66) \\
    + {GconAttn \scalebox{0.7}{\citep{wang2019improving}}}  & 47.95~($\pm$0.11)  & 54.96~($\pm$0.69) & 72.61($~\pm$0.39) &
    68.59~($\pm$0.96)\\
    + {KagNet \scalebox{0.7}{\citep{lin-etal-2019-kagnet}}}  & -- & -- & 73.47~($\pm$0.22) & 69.01~($\pm$0.76) \\
    + {RN} \scalebox{0.7}{\citep{santoro2017simple}}  & 45.12~($\pm$0.69)  & 54.23~($\pm$0.28) & 74.57~($\pm$0.91) & 69.08~($\pm$0.21) \\
    + {MHGRN} \scalebox{0.7}{\citep{feng-etal-2020-scalable}} & -- & -- & 74.45~($\pm$0.10)   & {71.11}~($\pm$0.81)  \\
    + {PGQA} \scalebox{0.7}{\citep{pgqa}} & 58.25~($\pm$0.43)  & \underline{69.66}~($\pm$0.97) & \underline{77.53}~($\pm$0.47)$^{q}$   & 71.19~($\pm$0.49)  \\
    + {QA-GNN} \scalebox{0.7}{\citep{Yasunaga2021QAGNN}} & \underline{59.08}~($\pm$1.25)  & 68.70~($\pm$0.62) &  75.54~($\pm$0.42)   & \underline{72.29}~($\pm$0.43)$^{p}$  \\
    \midrule
    + \pgsname\ (\textbf{Ours}) & \textbf{61.20}~($\pm$0.19)$^{p,q}$  & \textbf{70.23}~($\pm$0.40)$^{q}$ &  \textbf{78.15}~($\pm$0.23)$^{p,q}$   & \textbf{72.87}~($\pm$0.31)$^{p,q}$  \\
    \bottomrule 
\end{tabular}
}
% \vspace{-2mm}
\caption{Performance comparison on in-house dev (\textbf{IHdev}) and test (\textbf{IHtest}) split of \textbf{\textit{CommonsenseQA}} dataset. All scores are averaged across $5$ runs. First row depicts amount of training data used. The second-best number for each column is underlined while best is in bold. Superscripts `$p$' and `$q$' denote statistically significant differences (\textbf{\textit{p-value} of 0.05}) in comparison to two of our baselines- PGQA and QA-GNN, respectively.
}
% \vspace{-2mm}
\label{tab:csqa_inhouse}
\end{table*}
%%%

% Here the correct answer is `web page'

We perform multiple choice question answering on the CSQA dataset~\citep{talmor-etal-2019-commonsenseqa}. 
% which is most commonly used benchmark for this task. 
Here, a question is given with 5 answer choices and the model has to predict the correct one. As an example, consider a question `Where could you see an advertisement while reading news?' with answer choices `television, bus, email, web page, and la villa'. One of the prior works for this task - PGQA~\citep{pgqa}, comprises of a knowledge module which generates commonsense and a QA module which identifies correct choice using this knowledge (see \cref{app:pgqa_details} for details). Since our aim is not to design an improved QA module but a better commonsense generator, for fair comparison with PGQA, we use their QA module with \pgsname. The QA module embeds the question + choices using RoBERTa~\citep{roberta} and uses the CLS token output to perform attention over path embeddings generated using the commonsense module. The output of attention module together with embedding of question and answer choices is used to predict the correct answer. 
% We elaborate the details of PGQA in \cref{app:pgqa_details}.

% We just replace their paths with the ones generated by \pgsname

Table \ref{tab:csqa_inhouse} shows results on CSQA which are usually averaged over 5 runs on this benchmark. We compare against several baselines broadly classified into ones using static KG such as MHGRN~\citep{feng-etal-2020-scalable}, QA-GNN~\citep{Yasunaga2021QAGNN} etc. and others which train a dynamic path generator (PGQA)~\citep{pgqa} as commonsense module. We also compare with T5-base since it is backbone LM for \pgsname. When using entire training data, we observe that \pgsname\ performs better than all baselines\footnote{Results for \href{https://github.com/wangpf3/Commonsense-Path-Generator}{PGQA} and \href{https://github.com/michiyasunaga/qagnn}{QA-GNN} are reproduced using their official open-source implementations while numbers for other baselines have been taken from these two works.} 
% \footnote{The source code for baselines was taken from their official GitHub repositories.}
on test set. 
% while performs second best on the dev split. 
% PGQA~\citep{pgqa} is most similar to our approach since we adapt their framework and just differ in how commonsense paths are generated.
We outperform PGQA with a gain of $1.68\%$ in accuracy on test split signifying the relevance and applicability of inferences generated by \pgsname. \pgsname\ performs better than QA-GNN~\citep{Yasunaga2021QAGNN} also particularly in low training data regimes with performance gains of $\sim 2\%$ (and $\sim 3\%$ over PGQA) showing that while QA-GNN is more sensitive towards amount of training data used, \pgsname\ is more robust and helps in generalizing better. Qualitatively, consider the question  - ‘Where could you see an advertisement while reading news?’ PGQA generates the path - ‘read\_news  \textit{hassubevent}  read  \textit{relatedto}  news  \textit{atlocation}  television’ ignoring the context that advertisement is being seen along with reading news and ends up predicting television as answer which is wrong. While \pgsname\ generates - ‘spread\_information  \textit{\_capableof} advertisement  \textit{atlocation}  web\_page  \textit{usedfor}  reading\_news’. Here it can be seen that \pgsname\ identifies that seeing the advertisement and reading news is happening together and generates path accordingly to relate them with ‘web page’ which is the correct answer.
We also conduct a thorough qualitative comparison (\cref{ap:qcsqa}) where we observe that evaluators find \pgsname\ paths to be significantly more contextually relevant than PGQA. \\
\indent We conduct a \textbf{human study} wherein we presented evaluators with questions from CSQA dataset with corresponding commonsense paths generated by \pgsname\ and PGQA in an anonymized manner to compare the generative commonsense methods. We asked them to compare the paths based on their contextual relevance with the complete sentence and classify them into one of three categories - 1) ‘\pgsname\ is better than PGQA’, 2) ‘PGQA is better than \pgsname’, 3) ‘Both are of the similar quality’. A total of 150 questions samples were randomly sampled from the test set and presented to 6 evaluators (25 samples each). Following are our observations:

\begin{itemize}
    \item Number of samples where \pgsname\ is better: 62 (41.33\% of 150 samples) 
    \item Number of samples where PGQA is better: 38 (25.33\% of 150 samples) 
    \item Number of samples where both are of similar quality: 50 (33.33\% of 150 samples)
\end{itemize}

%% Common table for all ablations
\setlength{\belowcaptionskip}{-10pt}
\begin{table*}[t]
\centering
\scalebox{0.85}{
\begin{tabular}{cc|cc|cc}
\hline
\multicolumn{2}{c|}{\begin{tabular}[c]{@{}c@{}}Entity masking\\ for training\end{tabular}} &
  \multicolumn{2}{c|}{\begin{tabular}[c]{@{}c@{}}Query template\\ for path-sentence\end{tabular}} &
  \multicolumn{2}{c}{\begin{tabular}[c]{@{}c@{}}Entity masking\\ for inference\end{tabular}} \\ \hline
$p_{mask}$ &
  IHdev (\%) &
  Query &
  IHdev (\%) &
  Type &
  IHdev (\%) \\ \hline
0.0 &
  77.52~($\pm$0.44) &
  \multirow{2}{*}{$Q1$} &
  \multirow{2}{*}{77.69~($\pm$0.43)} &
  \multirow{2}{*}{\begin{tabular}[c]{@{}c@{}}\textit{Interrogative} \end{tabular}} &
  \multirow{2}{*}{78.07~($\pm$0.56)} \\
0.50 &
  77.38~($\pm$0.40) &
   &
   &
   &
   \\
0.67 &
  77.61~($\pm$0.79) &
  \multirow{2}{*}{$Q2$} &
  \multirow{2}{*}{77.25~($\pm$0.64)} &
  \multirow{2}{*}{\begin{tabular}[c]{@{}c@{}}\textit{Random} \end{tabular}} &
  \multirow{2}{*}{77.90~($\pm$0.84)} \\
1.0 &
  77.71~($\pm$1.17) &
   &
   &
   &
   \\ \hline
\textbf{0.33} &
  \textbf{78.15}~($\pm$0.23) &
  $\mathbf{Q1+Q2}$ &
  \textbf{78.15}~($\pm$0.23) &
  \textbf{No Masking} &
  \textbf{78.15}~($\pm$0.23) \\ \hline
\end{tabular}
}
\caption{Studying the effect of ablation variants through comparison on \textit{CommonsenseQA} dev set.
}
\label{tab:ablation}
\end{table*}
%%%

This shows that commonsense generated by \pgsname\ is found to be more relevant in human evaluation. Also, if we exclude neutral samples and consider the 100 samples where the commonsense paths generated by one of either approach is found to be better, \pgsname’s paths are found to be more relevant in 62 samples (62\% of 100 samples) while PGQA’s paths are more relevant in 38 samples (38\% of 100 samples).
% Please refer to \cref{ap:qcsqa} for human study and more qualitative comparison with baselines on CSQA dataset. 

We also study the effect of using a different generative LM (GPT-2 as used by PGQA) as backbone for \pgsname\ in \cref{app:gpt_experiments} and empirically establish that performance gains over PGQA are independent of which LM is used.

% To recognise the influence of our methodology over baseline methods

% Thus, we perform further experiments keeping $p_{mask}=0.33$ during training

\subsubsection{Ablation Study}
\label{sec:abl}
\paragraph{Entity masking during training}
As described in \cref{sec:s_to_p_pg}, 
a parameter
%probability 
$p_{mask}$ 
is used to decide 
% governs 
whether entities in an input sentence 
will be masked. 
% This enables \pgsname\ to capture context better by identifying masked entity. 
We tune $p_{mask}$ over the CSQA IHdev set and determine $0.33$ as optimal value. 
Table \ref{tab:ablation} shows comparison where masking during training works better than not masking. We show qualitative analysis for different $p_{mask}$ in \cref{ap:mask}. Further, $0 < p_{mask} < 1$ ensures trained \pgsname\ can be used for both masked and unmasked inputs.

% \vspace{-2mm}
\paragraph{Path-sentence query templates}
% We obtain path-sentence mapping in our dataset by devising two query templates $Q1$ (includes relation information) and $Q2$ (does not capture relations) as described in \cref{sec:s_to_p}. Table \ref{tab:ablation} shows a performance comparison of using different query templates. 
As described in \cref{sec:s_to_p}, we used two query templates---$Q1$ (includes relation information) and $Q2$ (does not capture relations)---while creating our path-sentence paired dataset.
Here we study the effect of using these different query templates (Table \ref{tab:ablation}).
% on CSQA dev set. 
% We observe that training \pgsname\ on dataset created using $Q1$ alone outperforms the one trained only on $Q2$, indicating the improvement due to relation information. 
We observe that training \pgsname\ on a combined dataset, $Q1+Q2$, results in the best performance, followed by that on using $Q1$ alone, that further outperforms $Q2$. 
This indicates the influence of including relation information in the training dataset.
% It is to be noted that both these variants of the dataset are of the same size. 
% We also observe that combining these two variants into a single dataset, $Q1+Q2$, results in further improvement.

% \vspace{-2mm}
\paragraph{Entity masking during inference}
Since \pgsname\ is given a masked sentence as input during training ($p_{mask}=0.33$), we explore the effect of similar type of masking during inference. Specifically, certain parts of input sentence can be replaced with masked token to enable \pgsname\ to generate paths that lead towards filling the mask.
% in order to identify a default base variant, we test the performance with different masking strategies during inference on the CSQA development set. 
As reported in Table \ref{tab:ablation}, the variant where no masking is done performs marginally better than when \textit{Interrogative} or \textit{Random} tokens in sentence are masked. Thus, by default we do not perform masking during inference unless otherwise stated. 

\subsection{OpenCSR: Open-Ended CommonSense Reasoning} \label{sec:down_opencsr}
% \vspace{-1mm}
In CSQA, often multiple choices are appropriate and model gets penalised unfairly if it predicts suitable answer which does not match with single ground truth. To mitigate this, \citet{lin-etal-2019-kagnet} re-configured three multi-choice QA datasets for OpenCSR as a generative task where interrogative tokens are replaced with blanks (``\_ \_'') and a set of singleton tokens is labeled as ground truth. To generate a set of paths $P$, we use inference masking variant of \pgsname\ since question contains a blank. Given a question $q$, blank (``\_ \_'') is replaced with mask token. To inject our paths, we devise a supervised method where we adapt a separate T5-base model for OpenCSR such that concatenation of $q$ and paths is given as input to T5 along with the prefix `\textit{fill mask to answer question: }'. T5 is trained to generate one of the answers in ground truth set. During inference, top-$K$ answers, determined on basis of generation likelihood from T5 decoder, are taken as answer candidates.

% \vspace{-1mm}
Table \ref{tab:opencsr_eval} shows comparison between DrFact\footnote{The authors communicated that the test set and leader board has not been released yet. Hence, we report results using the author provided \href{https://github.com/yuchenlin/OpenCSR/}{code and validation set}. Also, they run their models on single seed.}~\citep{lin-etal-2021-differentiable} (current state-of-the-art) and our supervised method which uses \pgsname's paths. Specifically, we evaluate - 1) `Paths from \pgsname' where generated paths are concatenated; and 2) `Concepts from \pgsname' where only entities in generated paths are appended. Since our supervised method is based on pre-trained T5, for fair comparison and to probe if performance changes are due to T5, we compare against another baseline: T5-base fine tuned for OpenCSR without paths. We evaluate two metrics as used in \citet{lin-etal-2021-differentiable}: 1) \textbf{Hits@K}: Determined on basis of whether generated and ground truth answer sets have non-empty intersection; 2) \textbf{Recall@K}: Estimates how many predicted answers match at least one ground truth answer. We vary value of K to be \{10, 30, 50\}. We evaluate on three datasets - ARC~\citep{clark2018think}, QASC~\citep{Khot_Clark_Guerquin_Jansen_Sabharwal_2020}, and OBQA~\citep{obqa}. 

%% OpenCSR Results
% \setlength{\belowcaptionskip}{-10pt}
\begin{table*}[t]
\centering
% \resizebox{\textwidth}{!}{%
\scalebox{0.92}{
\begin{tabular}{ll|ccc|ccc|ccc}
\hline
\multicolumn{1}{l}{} &
   &
  \multicolumn{3}{c|}{\textbf{ARC}} &
  \multicolumn{3}{c|}{\textbf{QASC}} &
  \multicolumn{3}{c}{\textbf{OBQA}} \\ 
  \hline
\multicolumn{2}{c|}{\textbf{Hits@K}} &
  {H@10} &
  {H@30} &
  {H@50} &
  {H@10} &
  {H@30} &
  {H@50} &
  {H@10} &
  {H@30} &
  {H@50} \\ \hline
\multicolumn{2}{l|}{{DrFact \scalebox{0.7}{\citep{lin-etal-2021-differentiable}}}} &
  {36.09} &
  {53.25} &
  {64.50} &
  {21.78} &
  {37.62} &
  {51.49} &
  {12.08} &
  {23.77} &
  {35.13} \\
\multicolumn{2}{l|}{{T5-base \scalebox{0.7}{\citep{raffel2019t5}}}} &
  {\underline{49.70}} &
  {\textbf{67.46}} &
  {\underline{71.01}} &
  {\underline{33.66}} &
  {\underline{47.52}} &
  {53.47} &
  {17.42} &
  {29.55} &
  {37.88} \\
\multicolumn{2}{l|}{+ {\pgsname\ Paths }} &
  {\textbf{50.89}} &
  {63.91} &
  {69.23} &
  {30.69} &
  {\underline{47.52}} &
  {\underline{56.44}} &
  {\underline{20.45}} &
  {\underline{34.09}} &
  {\textbf{45.45}} \\
\multicolumn{2}{l|}{+ {\pgsname\ Concepts }} &
  {44.97} &
  {\underline{66.86}} &
  {\textbf{73.37}} &
  {\textbf{35.64}} &
  {\textbf{47.52}} &
  {\textbf{57.43}} &
  {\textbf{21.21}} &
  {\textbf{35.61}} &
  {\underline{42.42}} \\ \hline
\multicolumn{2}{c|}{\textbf{Recall@K}} &
  {R@10} &
  {R@30} &
  {R@50} &
  {R@10} &
  {R@30} &
  {R@50} &
  {R@10} &
  {R@30} &
  {R@50} \\ \hline
\multicolumn{2}{l|}{{DrFact \scalebox{0.7}{\citep{lin-etal-2021-differentiable}}}} &
  {12.60} &
  \multicolumn{1}{c}{{21.05}} &
  {27.27} &
  {12.38} &
  {22.28} &
  {29.70} &
  {6.12} &
  {11.85} &
  {16.51} \\
\multicolumn{2}{l|}{{T5-base \scalebox{0.7}{\citep{raffel2019t5}}}} &
  {\underline{15.98}} &
  \multicolumn{1}{c}{{\underline{28.30}}} &
  {\underline{33.93}} &
  {\underline{18.98}} &
  {26.40} &
  {30.53} &
  {8.52} &
  {14.61} &
  {18.71} \\
\multicolumn{2}{l|}{+ {\pgsname\ Paths }} &
  {\textbf{16.87}} &
  \multicolumn{1}{c}{{27.45}} &
  {33.73} &
  {17.49} &
  {\underline{28.05}} &
  {\textbf{33.33}} &
  {\underline{9.90}} &
  {\underline{16.53}} &
  {\textbf{22.42}} \\
\multicolumn{2}{l|}{+ {\pgsname\ Concepts }} &
  {15.12} &
  \multicolumn{1}{c}{{\textbf{28.99}}} &
  {\textbf{35.21}} &
  {\textbf{19.64}} &
  {\textbf{28.05}} &
  {\underline{33.00}} &
  {\textbf{9.96}} &
  {\textbf{17.35}} &
  {\underline{21.10}} \\ \hline
\end{tabular}%
}

\caption{Performance comparison on Hits@K and Recall@K metrics for OpenCSR~\citep{lin-etal-2021-differentiable} on ARC, QASC and OBQA datasets. DrFact is a BERT-based current state of the art method.}
\label{tab:opencsr_eval}
\end{table*}
%%%

% \vspace{-1mm}
\pgsname\ performs significantly better than both baselines on all datasets uniformly. Specifically, `Concepts from \pgsname' usually performs better which shows entities in paths generated by \pgsname\ are useful. Our approach provides performance gains of upto $8\%$, $6\%$, $10\%$ in Hits@50 and $8\%$, $3\%$, $6\%$ in Recall@50 over DrFact on ARC, QASC and OBQA respectively. Even though T5-base baseline performs better than DrFact, commonsense from \pgsname\ augmented with T5 achieves new state of the art on this task with performance gains upto $2.3\%$, $3.9\%$, $7.5\%$ in Hits@50 and $1.2\%$, $2.5\%$, $3.7\%$ in Recall@50 over T5-base on ARC, QASC and OBQA respectively.

% \vspace{-1mm}
\subsection{Effect of Concatenating \pgsname\ Knowledge in Generation Task}
\label{sec:other_tasks}
We explore augmenting \pgsname\ paths for text generation where our aim is not to obtain SOTA results but to analyse if it improves performance of a base model. Specifically, we study Paraphrase Generation: given a sentence, generate another sentence expressing same meaning using different words where commonsense is usually needed while rephrasing. Since T5~\citep{raffel2019t5} is designed for generation tasks, we fine-tune T5-base to generate annotated paraphrase given a sentence as input on MRPC dataset~\citep{dolan-brockett-2005-automatically}. Generated paths are appended as string to input. Please refer to appendix \ref{sec:other_tasks_app} for elaborated implementation details and discussion. 

\begin{table}[h]
\centering
\scalebox{0.48}{
\resizebox{\textwidth}{!}{%
\begin{tabular}{lccccc}
\hline
\multicolumn{6}{c}{\textbf{MRPC Paraphrase Generation}}     \\ 
\hline
                                & BLEU-4 & METEOR & ROUGE-L & CIDEr & SPICE \\ 
                                % \hline
T5-base                         & 43.10  & 36.10  & 61.80   & 36.33 & 47.10  \\
+ \pgsname\ Paths & \textbf{44.50}  & \textbf{36.70}  & \textbf{62.50}   & \textbf{37.34} & \textbf{48.50}   \\ \cline{1-6}
% \multicolumn{6}{c}{\textbf{CommonGen}}        \\ \cline{1-6}
% T5-base                         & 9.90   & 21.10  & \textbf{36.70}   & 14.54 & \textbf{44.70}   \\
% + PGQA Paths                         & 9.80   & 21.0  & 36.60   & 14.64 & \textbf{44.70}   \\
% + \pgsname\ Paths & \textbf{10.10}  & \textbf{21.20}  & \textbf{36.70}   & \textbf{14.78} & \textbf{44.70}   \\ \hline
\end{tabular}}
}
\caption{Using \pgsname\ Paths leads to improvements in paraphrase generation task on MRPC dataset. Generative commonsense methods like PGQA which rely on answer choices cannot be applied in tasks like paraphrase generation where entities are not available.
% Please note that performance might not match with leaderboard numbers and other tuned implementations since this is our implementation and we did not tune the hyper-parameters while performing these experiments. But we used the same configuration uniformly for both with and without commonsense path concatenation experiments.
}
\label{tab:generative_main_paper}
\end{table}

% \vspace{-1mm}
Table \ref{tab:generative_main_paper} summarises results evaluated through commonly used generation metrics - BLEU~\citep{papineni2002bleu}, METEOR~\citep{banerjee2005meteor}, ROUGE-L~\citep{lin2004rouge}, CIDEr~\citep{vedantam2015cider} and SPICE~\citep{anderson2016spice}. Amongst these, SPICE is considered to correlate most with human judgement. Using \pgsname\ paths results in better paraphrase generation as indicated by $\sim$1-1.5\% improvement in most metrics. 

% We would like to highlight that generative commonsense methods like PGQA which rely on answer choices cannot be applied in tasks like paraphrase generation where entities are not available.

% For CommonGen, no significant improvements are observed and performance remains similar without any degradation.
% \input{files/evaluation}
% \input{files/ablation}
% \vspace{-3mm}
\section{Conclusion}
% \vspace{-2mm}
We presented \pgsname, a framework to generate commonsense inferences that are relevant to the overall context of a given natural language text.
We created a novel dataset of $<$sentence, commonsense paths$>$ pairs for training \pgsname\ and make it available to the community\footnote{An ethics statement regarding the released dataset has been discussed in Appendix~\ref{ap:ethics}.}. Empirical evaluation shows that commonsense inferences generated by \pgsname\ are relevant, diverse, and also contain novel entities not present in KG. We augment knowledge generated by \pgsname\ in commonsense tasks such as Multi-Choice QA and Open-ended CommonSense Reasoning, achieving SoTA results for these tasks. Further, we also used \pgsname\ for NLP tasks such as paraphrase generation achieving improved performance. While, using ConceptNet as our base KG allowed us to perform an exhaustive fair comparison with a variety of benchmark methods---where the motivation is to provide more relevant knowledge (in symbolic form as in KG) to tasks---\pgsname\ can further be enhanced by utilizing other commonsense KGs. Our work can be extended to explore better ways of integrating the generated knowledge generically across a variety of KGs and LMs, and is a potential direction for future work.

% Also, new ways to utilize the  generated knowledge and augment commonsense paths in downstream tasks can be explored.

% \section*{Reproducibility Statement}
% We devise a methodology to create a novel sentence-commonsense paths data to train \pgsname. We release the dataset and trained \pgsname\ model \href{https://drive.google.com/drive/folders/19kGz-fGkYM-FAtcT47QiaUclpNd7ZAXC?usp=sharing}{here}. We provide detailed steps of our algorithm to create the dataset and training \pgsname\ model (\cref{sec:methodology}) accompanied with the detailed implementation details (\cref{sec:imp_details}). For each downstream task where \pgsname\ is used to augment knowledge, we provide implementation details and datasets used in the main paper (\cref{sec:analyse_p}, \cref{sec:down_csqa}, \cref{sec:down_opencsr} and \cref{sec:other_tasks}) as well as further in-depth details in \cref{ap:imp_det}. Links to sources of baseline codes and different datasets used for each task are also provided in these sections.

% Unlike previous generative approaches, which ignore important contextual clues in the input text and operate on entities, \pgsname\ works directly on input sentences making it suitable for a wide variety of NLP tasks

\bibliography{acl_latex}

\begin{thebibliography}{59}
\expandafter\ifx\csname natexlab\endcsname\relax\def\natexlab#1{#1}\fi

\bibitem[{Anderson et~al.(2016)Anderson, Fernando, Johnson, and
  Gould}]{anderson2016spice}
Peter Anderson, Basura Fernando, Mark Johnson, and Stephen Gould. 2016.
\newblock Spice: Semantic propositional image caption evaluation.
\newblock In \emph{European conference on computer vision}, pages 382--398.
  Springer.

\bibitem[{Banerjee and Lavie(2005)}]{banerjee2005meteor}
Satanjeev Banerjee and Alon Lavie. 2005.
\newblock Meteor: An automatic metric for mt evaluation with improved
  correlation with human judgments.
\newblock In \emph{Proceedings of the acl workshop on intrinsic and extrinsic
  evaluation measures for machine translation and/or summarization}, pages
  65--72.

\bibitem[{Bao et~al.(2016)Bao, Duan, Yan, Zhou, and
  Zhao}]{bao-etal-2016-constraint}
Junwei Bao, Nan Duan, Zhao Yan, Ming Zhou, and Tiejun Zhao. 2016.
\newblock \href {https://aclanthology.org/C16-1236} {Constraint-based question
  answering with knowledge graph}.
\newblock In \emph{Proceedings of {COLING} 2016, the 26th International
  Conference on Computational Linguistics: Technical Papers}, pages 2503--2514,
  Osaka, Japan. The COLING 2016 Organizing Committee.

\bibitem[{Becker et~al.(2021)Becker, Korfhage, Paul, and
  Frank}]{becker-EtAl:2021:IWCS}
Maria Becker, Katharina Korfhage, Debjit Paul, and Anette Frank. 2021.
\newblock \href {https://www.aclweb.org/anthology/2021.iwcs-1.3} {Co-nnect: A
  framework for revealing commonsense knowledge paths as explicitations of
  implicit knowledge in texts}.
\newblock In \emph{Proceedings of the 14th International Conference on
  Computational Semantics (IWCS)}, pages 21--32, Groningen, The Netherlands
  (online). Association for Computational Linguistics.

\bibitem[{Bhagavatula et~al.(2020)Bhagavatula, Bras, Malaviya, Sakaguchi,
  Holtzman, Rashkin, Downey, tau Yih, and Choi}]{bhagavatula2019abductive}
Chandra Bhagavatula, Ronan~Le Bras, Chaitanya Malaviya, Keisuke Sakaguchi, Ari
  Holtzman, Hannah Rashkin, Doug Downey, Wen tau Yih, and Yejin Choi. 2020.
\newblock \href {https://openreview.net/forum?id=Byg1v1HKDB} {Abductive
  commonsense reasoning}.
\newblock In \emph{International Conference on Learning Representations}.

\bibitem[{Bordes et~al.(2013)Bordes, Usunier, Garcia-Duran, Weston, and
  Yakhnenko}]{bordes2013translating}
Antoine Bordes, Nicolas Usunier, Alberto Garcia-Duran, Jason Weston, and Oksana
  Yakhnenko. 2013.
\newblock Translating embeddings for modeling multi-relational data.
\newblock \emph{Advances in neural information processing systems}, 26.

\bibitem[{Bosselut et~al.(2021)Bosselut, Bras, and Choi}]{bosselut2020dynamic}
Antoine Bosselut, Ronan~Le Bras, and Yejin Choi. 2021.
\newblock \href {https://ojs.aaai.org/index.php/AAAI/article/view/16625}
  {Dynamic neuro-symbolic knowledge graph construction for zero-shot
  commonsense question answering}.
\newblock In \emph{Thirty-Fifth {AAAI} Conference on Artificial Intelligence,
  {AAAI} 2021, Thirty-Third Conference on Innovative Applications of Artificial
  Intelligence, {IAAI} 2021, The Eleventh Symposium on Educational Advances in
  Artificial Intelligence, {EAAI} 2021, Virtual Event, February 2-9, 2021},
  pages 4923--4931. {AAAI} Press.

\bibitem[{Bosselut et~al.(2019)Bosselut, Rashkin, Sap, Malaviya, Celikyilmaz,
  and Choi}]{comet}
Antoine Bosselut, Hannah Rashkin, Maarten Sap, Chaitanya Malaviya, Asli
  Celikyilmaz, and Yejin Choi. 2019.
\newblock \href {https://doi.org/10.18653/v1/P19-1470} {{COMET}: Commonsense
  transformers for automatic knowledge graph construction}.
\newblock In \emph{Proceedings of the 57th Annual Meeting of the Association
  for Computational Linguistics}, pages 4762--4779, Florence, Italy.
  Association for Computational Linguistics.

\bibitem[{Clark et~al.(2018)Clark, Cowhey, Etzioni, Khot, Sabharwal, Schoenick,
  and Tafjord}]{clark2018think}
Peter Clark, Isaac Cowhey, Oren Etzioni, Tushar Khot, Ashish Sabharwal, Carissa
  Schoenick, and Oyvind Tafjord. 2018.
\newblock Think you have solved question answering? try arc, the ai2 reasoning
  challenge.
\newblock \emph{arXiv preprint arXiv:1803.05457}.

\bibitem[{Devlin et~al.(2019)Devlin, Chang, Lee, and
  Toutanova}]{devlin-etal-2019-bert}
Jacob Devlin, Ming-Wei Chang, Kenton Lee, and Kristina Toutanova. 2019.
\newblock \href {https://doi.org/10.18653/v1/N19-1423} {{BERT}: Pre-training of
  deep bidirectional transformers for language understanding}.
\newblock In \emph{Proceedings of the 2019 Conference of the North {A}merican
  Chapter of the Association for Computational Linguistics: Human Language
  Technologies, Volume 1 (Long and Short Papers)}, pages 4171--4186,
  Minneapolis, Minnesota. Association for Computational Linguistics.

\bibitem[{Dolan and Brockett(2005)}]{dolan-brockett-2005-automatically}
William~B. Dolan and Chris Brockett. 2005.
\newblock \href {https://aclanthology.org/I05-5002} {Automatically constructing
  a corpus of sentential paraphrases}.
\newblock In \emph{Proceedings of the Third International Workshop on
  Paraphrasing ({IWP}2005)}.

\bibitem[{Feng et~al.(2020)Feng, Chen, Lin, Wang, Yan, and
  Ren}]{feng-etal-2020-scalable}
Yanlin Feng, Xinyue Chen, Bill~Yuchen Lin, Peifeng Wang, Jun Yan, and Xiang
  Ren. 2020.
\newblock \href {https://doi.org/10.18653/v1/2020.emnlp-main.99} {Scalable
  multi-hop relational reasoning for knowledge-aware question answering}.
\newblock In \emph{Proceedings of the 2020 Conference on Empirical Methods in
  Natural Language Processing (EMNLP)}, pages 1295--1309, Online. Association
  for Computational Linguistics.

\bibitem[{Guu et~al.(2015)Guu, Miller, and Liang}]{guu-etal-2015-traversing}
Kelvin Guu, John Miller, and Percy Liang. 2015.
\newblock \href {https://doi.org/10.18653/v1/D15-1038} {Traversing knowledge
  graphs in vector space}.
\newblock In \emph{Proceedings of the 2015 Conference on Empirical Methods in
  Natural Language Processing}, pages 318--327, Lisbon, Portugal. Association
  for Computational Linguistics.

\bibitem[{Huang et~al.(2019)Huang, Le~Bras, Bhagavatula, and
  Choi}]{huang-etal-2019-cosmos}
Lifu Huang, Ronan Le~Bras, Chandra Bhagavatula, and Yejin Choi. 2019.
\newblock \href {https://doi.org/10.18653/v1/D19-1243} {Cosmos {QA}: Machine
  reading comprehension with contextual commonsense reasoning}.
\newblock In \emph{Proceedings of the 2019 Conference on Empirical Methods in
  Natural Language Processing and the 9th International Joint Conference on
  Natural Language Processing (EMNLP-IJCNLP)}, pages 2391--2401, Hong Kong,
  China. Association for Computational Linguistics.

\bibitem[{Ilievski et~al.(2021)Ilievski, Szekely, and Zhang}]{ilievski2021cskg}
Filip Ilievski, Pedro Szekely, and Bin Zhang. 2021.
\newblock Cskg: The commonsense knowledge graph.
\newblock In \emph{European Semantic Web Conference}, pages 680--696. Springer.

\bibitem[{Jiang et~al.(2020)Jiang, Xu, Araki, and Neubig}]{whatLMKnow}
Zhengbao Jiang, Frank~F Xu, Jun Araki, and Graham Neubig. 2020.
\newblock \href {https://www.aclweb.org/anthology/2020.tacl-1.28.pdf} {How can
  we know what language models know?}
\newblock \emph{Transactions of the Association for Computational Linguistics},
  8:423--438.

\bibitem[{Kassner and Sch{\"u}tze(2020)}]{kassner-schutze-2020-negated}
Nora Kassner and Hinrich Sch{\"u}tze. 2020.
\newblock \href {https://doi.org/10.18653/v1/2020.acl-main.698} {Negated and
  misprimed probes for pretrained language models: Birds can talk, but cannot
  fly}.
\newblock In \emph{Proceedings of the 58th Annual Meeting of the Association
  for Computational Linguistics}, pages 7811--7818, Online. Association for
  Computational Linguistics.

\bibitem[{Khot et~al.(2020)Khot, Clark, Guerquin, Jansen, and
  Sabharwal}]{Khot_Clark_Guerquin_Jansen_Sabharwal_2020}
Tushar Khot, Peter Clark, Michal Guerquin, Peter Jansen, and Ashish Sabharwal.
  2020.
\newblock \href {https://doi.org/10.1609/aaai.v34i05.6319} {Qasc: A dataset for
  question answering via sentence composition}.
\newblock \emph{Proceedings of the AAAI Conference on Artificial Intelligence},
  34(05):8082--8090.

\bibitem[{Li et~al.(2021)Li, Nye, and Andreas}]{li-etal-2021-implicit}
Belinda~Z. Li, Maxwell Nye, and Jacob Andreas. 2021.
\newblock \href {https://doi.org/10.18653/v1/2021.acl-long.143} {Implicit
  representations of meaning in neural language models}.
\newblock In \emph{Proceedings of the 59th Annual Meeting of the Association
  for Computational Linguistics and the 11th International Joint Conference on
  Natural Language Processing (Volume 1: Long Papers)}, pages 1813--1827,
  Online. Association for Computational Linguistics.

\bibitem[{Li et~al.(2016)Li, Taheri, Tu, and Gimpel}]{li-etal-2016-commonsense}
Xiang Li, Aynaz Taheri, Lifu Tu, and Kevin Gimpel. 2016.
\newblock \href {https://doi.org/10.18653/v1/P16-1137} {Commonsense knowledge
  base completion}.
\newblock In \emph{Proceedings of the 54th Annual Meeting of the Association
  for Computational Linguistics (Volume 1: Long Papers)}, pages 1445--1455,
  Berlin, Germany. Association for Computational Linguistics.

\bibitem[{Lin et~al.(2019)Lin, Chen, Chen, and Ren}]{lin-etal-2019-kagnet}
Bill~Yuchen Lin, Xinyue Chen, Jamin Chen, and Xiang Ren. 2019.
\newblock \href {https://doi.org/10.18653/v1/D19-1282} {{K}ag{N}et:
  Knowledge-aware graph networks for commonsense reasoning}.
\newblock In \emph{Proceedings of the 2019 Conference on Empirical Methods in
  Natural Language Processing and the 9th International Joint Conference on
  Natural Language Processing (EMNLP-IJCNLP)}, pages 2829--2839, Hong Kong,
  China. Association for Computational Linguistics.

\bibitem[{Lin et~al.(2021{\natexlab{a}})Lin, Sun, Dhingra, Zaheer, Ren, and
  Cohen}]{lin-etal-2021-differentiable}
Bill~Yuchen Lin, Haitian Sun, Bhuwan Dhingra, Manzil Zaheer, Xiang Ren, and
  William Cohen. 2021{\natexlab{a}}.
\newblock \href {https://www.aclweb.org/anthology/2021.naacl-main.366}
  {Differentiable open-ended commonsense reasoning}.
\newblock In \emph{Proceedings of the 2021 Conference of the North American
  Chapter of the Association for Computational Linguistics: Human Language
  Technologies}, pages 4611--4625, Online. Association for Computational
  Linguistics.

\bibitem[{Lin et~al.(2021{\natexlab{b}})Lin, Wu, Yang, Lee, and
  Ren}]{lin2021riddlesense}
Bill~Yuchen Lin, Ziyi Wu, Yichi Yang, Dong-Ho Lee, and Xiang Ren.
  2021{\natexlab{b}}.
\newblock Riddlesense: Reasoning about riddle questions featuring linguistic
  creativity and commonsense knowledge.
\newblock In \emph{Findings of the Association for Computational Linguistics:
  ACL-IJCNLP 2021}, pages 1504--1515.

\bibitem[{Lin et~al.(2020)Lin, Zhou, Shen, Zhou, Bhagavatula, Choi, and
  Ren}]{lin-etal-2020-commongen}
Bill~Yuchen Lin, Wangchunshu Zhou, Ming Shen, Pei Zhou, Chandra Bhagavatula,
  Yejin Choi, and Xiang Ren. 2020.
\newblock \href {https://doi.org/10.18653/v1/2020.findings-emnlp.165}
  {{C}ommon{G}en: A constrained text generation challenge for generative
  commonsense reasoning}.
\newblock In \emph{Findings of the Association for Computational Linguistics:
  EMNLP 2020}, pages 1823--1840, Online. Association for Computational
  Linguistics.

\bibitem[{Lin(2004)}]{lin2004rouge}
Chin-Yew Lin. 2004.
\newblock Rouge: A package for automatic evaluation of summaries.
\newblock In \emph{Text summarization branches out}, pages 74--81.

\bibitem[{Liu et~al.(2019)Liu, Ott, Goyal, Du, Joshi, Chen, Levy, Lewis,
  Zettlemoyer, and Stoyanov}]{roberta}
Yinhan Liu, Myle Ott, Naman Goyal, Jingfei Du, Mandar Joshi, Danqi Chen, Omer
  Levy, Mike Lewis, Luke Zettlemoyer, and Veselin Stoyanov. 2019.
\newblock \href {http://arxiv.org/abs/1907.11692} {Roberta: {A} robustly
  optimized {BERT} pretraining approach}.
\newblock \emph{CoRR}, abs/1907.11692.

\bibitem[{Loshchilov and Hutter(2017)}]{adamw2017}
Ilya Loshchilov and Frank Hutter. 2017.
\newblock \href {http://arxiv.org/abs/1711.05101} {Fixing weight decay
  regularization in adam}.
\newblock \emph{CoRR}, abs/1711.05101.

\bibitem[{Lv et~al.(2020)Lv, Guo, Xu, Tang, Duan, Gong, Shou, Jiang, Cao, and
  Hu}]{Lv_Guo_Xu_Tang_Duan_Gong_Shou_Jiang_Cao_Hu_2020}
Shangwen Lv, Daya Guo, Jingjing Xu, Duyu Tang, Nan Duan, Ming Gong, Linjun
  Shou, Daxin Jiang, Guihong Cao, and Songlin Hu. 2020.
\newblock \href {https://doi.org/10.1609/aaai.v34i05.6364} {Graph-based
  reasoning over heterogeneous external knowledge for commonsense question
  answering}.
\newblock \emph{Proceedings of the AAAI Conference on Artificial Intelligence},
  34(05):8449--8456.

\bibitem[{Mihaylov et~al.(2018)Mihaylov, Clark, Khot, and Sabharwal}]{obqa}
Todor Mihaylov, Peter Clark, Tushar Khot, and Ashish Sabharwal. 2018.
\newblock \href {https://doi.org/10.18653/v1/D18-1260} {Can a suit of armor
  conduct electricity? a new dataset for open book question answering}.
\newblock In \emph{Proceedings of the 2018 Conference on Empirical Methods in
  Natural Language Processing}, pages 2381--2391, Brussels, Belgium.
  Association for Computational Linguistics.

\bibitem[{Mitra et~al.(2019)Mitra, Banerjee, Pal, Mishra, and
  Baral}]{mitra2019exploring}
Arindam Mitra, Pratyay Banerjee, Kuntal~Kumar Pal, Swaroop Mishra, and Chitta
  Baral. 2019.
\newblock Exploring ways to incorporate additional knowledge to improve natural
  language commonsense question answering.
\newblock \emph{arXiv preprint arXiv:1909.08855}.

\bibitem[{Niven and Kao(2019)}]{niven-kao-2019-probing}
Timothy Niven and Hung-Yu Kao. 2019.
\newblock \href {https://doi.org/10.18653/v1/P19-1459} {Probing neural network
  comprehension of natural language arguments}.
\newblock In \emph{Proceedings of the 57th Annual Meeting of the Association
  for Computational Linguistics}, pages 4658--4664, Florence, Italy.
  Association for Computational Linguistics.

\bibitem[{Papineni et~al.(2002)Papineni, Roukos, Ward, and
  Zhu}]{papineni2002bleu}
Kishore Papineni, Salim Roukos, Todd Ward, and Wei-Jing Zhu. 2002.
\newblock Bleu: a method for automatic evaluation of machine translation.
\newblock In \emph{Proceedings of the 40th annual meeting of the Association
  for Computational Linguistics}, pages 311--318.

\bibitem[{Petroni et~al.(2019)Petroni, Rockt{\"a}schel, Riedel, Lewis, Bakhtin,
  Wu, and Miller}]{lmaskb}
Fabio Petroni, Tim Rockt{\"a}schel, Sebastian Riedel, Patrick Lewis, Anton
  Bakhtin, Yuxiang Wu, and Alexander Miller. 2019.
\newblock \href {https://doi.org/10.18653/v1/D19-1250} {Language models as
  knowledge bases?}
\newblock In \emph{Proceedings of the 2019 Conference on Empirical Methods in
  Natural Language Processing and the 9th International Joint Conference on
  Natural Language Processing (EMNLP-IJCNLP)}, pages 2463--2473, Hong Kong,
  China. Association for Computational Linguistics.

\bibitem[{Radford et~al.(2018{\natexlab{a}})Radford, Narasimhan, Salimans, and
  Sutskever}]{radford2018improving}
Alec Radford, Karthik Narasimhan, Tim Salimans, and Ilya Sutskever.
  2018{\natexlab{a}}.
\newblock Improving language understanding by generative pre-training.

\bibitem[{Radford et~al.(2018{\natexlab{b}})Radford, Narasimhan, Salimans, and
  Sutskever}]{gpt}
Alec Radford, Karthik Narasimhan, Tim Salimans, and Ilya Sutskever.
  2018{\natexlab{b}}.
\newblock Improving language understanding by generative pre-training.

\bibitem[{Raffel et~al.(2019)Raffel, Shazeer, Roberts, Lee, Narang, Matena,
  Zhou, Li, and Liu}]{raffel2019t5}
Colin Raffel, Noam Shazeer, Adam Roberts, Katherine Lee, Sharan Narang, Michael
  Matena, Yanqi Zhou, Wei Li, and Peter~J. Liu. 2019.
\newblock \href {http://arxiv.org/abs/1910.10683} {Exploring the limits of
  transfer learning with a unified text-to-text transformer}.
\newblock \emph{CoRR}, abs/1910.10683.

\bibitem[{Reimers and Gurevych(2019)}]{sbert}
Nils Reimers and Iryna Gurevych. 2019.
\newblock \href {https://doi.org/10.18653/v1/D19-1410} {Sentence-{BERT}:
  Sentence embeddings using {S}iamese {BERT}-networks}.
\newblock In \emph{Proceedings of the 2019 Conference on Empirical Methods in
  Natural Language Processing and the 9th International Joint Conference on
  Natural Language Processing (EMNLP-IJCNLP)}, pages 3982--3992, Hong Kong,
  China. Association for Computational Linguistics.

\bibitem[{Ren* et~al.(2020)Ren*, Hu*, and Leskovec}]{Ren*2020Query2box:}
Hongyu Ren*, Weihua Hu*, and Jure Leskovec. 2020.
\newblock \href {https://openreview.net/forum?id=BJgr4kSFDS} {Query2box:
  Reasoning over knowledge graphs in vector space using box embeddings}.
\newblock In \emph{International Conference on Learning Representations}.

\bibitem[{Ren and Leskovec(2020)}]{NEURIPS2020_e43739bb}
Hongyu Ren and Jure Leskovec. 2020.
\newblock \href
  {https://proceedings.neurips.cc/paper/2020/file/e43739bba7cdb577e9e3e4e42447f5a5-Paper.pdf}
  {Beta embeddings for multi-hop logical reasoning in knowledge graphs}.
\newblock In \emph{Advances in Neural Information Processing Systems},
  volume~33, pages 19716--19726. Curran Associates, Inc.

\bibitem[{Roberts et~al.(2020)Roberts, Raffel, and Shazeer}]{knowledgePackLM}
Adam Roberts, Colin Raffel, and Noam Shazeer. 2020.
\newblock \href {https://doi.org/10.18653/v1/2020.emnlp-main.437} {How much
  knowledge can you pack into the parameters of a language model?}
\newblock In \emph{Proceedings of the 2020 Conference on Empirical Methods in
  Natural Language Processing (EMNLP)}, pages 5418--5426, Online. Association
  for Computational Linguistics.

\bibitem[{Santoro et~al.(2017)Santoro, Raposo, Barrett, Malinowski, Pascanu,
  Battaglia, and Lillicrap}]{santoro2017simple}
Adam Santoro, David Raposo, David~G Barrett, Mateusz Malinowski, Razvan
  Pascanu, Peter Battaglia, and Timothy Lillicrap. 2017.
\newblock \href
  {https://proceedings.neurips.cc/paper/2017/file/e6acf4b0f69f6f6e60e9a815938aa1ff-Paper.pdf}
  {A simple neural network module for relational reasoning}.
\newblock In \emph{Advances in Neural Information Processing Systems},
  volume~30. Curran Associates, Inc.

\bibitem[{Sap et~al.(2019{\natexlab{a}})Sap, Le~Bras, Allaway, Bhagavatula,
  Lourie, Rashkin, Roof, Smith, and Choi}]{atomic}
Maarten Sap, Ronan Le~Bras, Emily Allaway, Chandra Bhagavatula, Nicholas
  Lourie, Hannah Rashkin, Brendan Roof, Noah~A Smith, and Yejin Choi.
  2019{\natexlab{a}}.
\newblock Atomic: An atlas of machine commonsense for if-then reasoning.
\newblock In \emph{Proceedings of the AAAI Conference on Artificial
  Intelligence}, volume~33, pages 3027--3035.

\bibitem[{Sap et~al.(2019{\natexlab{b}})Sap, Rashkin, Chen, Le~Bras, and
  Choi}]{sap-etal-2019-social}
Maarten Sap, Hannah Rashkin, Derek Chen, Ronan Le~Bras, and Yejin Choi.
  2019{\natexlab{b}}.
\newblock \href {https://doi.org/10.18653/v1/D19-1454} {Social {IQ}a:
  Commonsense reasoning about social interactions}.
\newblock In \emph{Proceedings of the 2019 Conference on Empirical Methods in
  Natural Language Processing and the 9th International Joint Conference on
  Natural Language Processing (EMNLP-IJCNLP)}, pages 4463--4473, Hong Kong,
  China. Association for Computational Linguistics.

\bibitem[{Schlichtkrull et~al.(2018)Schlichtkrull, Kipf, Bloem, Van Den~Berg,
  Titov, and Welling}]{schlichtkrull2018modeling}
Michael Schlichtkrull, Thomas~N Kipf, Peter Bloem, Rianne Van Den~Berg, Ivan
  Titov, and Max Welling. 2018.
\newblock Modeling relational data with graph convolutional networks.
\newblock In \emph{European semantic web conference}, pages 593--607. Springer.

\bibitem[{Speer et~al.(2017)Speer, Chin, and Havasi}]{conceptnet}
Robyn Speer, Joshua Chin, and Catherine Havasi. 2017.
\newblock Conceptnet 5.5: An open multilingual graph of general knowledge.
\newblock In \emph{Proceedings of the AAAI Conference on Artificial
  Intelligence}, volume~31.

\bibitem[{Sun et~al.(2018)Sun, Dhingra, Zaheer, Mazaitis, Salakhutdinov, and
  Cohen}]{sun-etal-2018-open}
Haitian Sun, Bhuwan Dhingra, Manzil Zaheer, Kathryn Mazaitis, Ruslan
  Salakhutdinov, and William Cohen. 2018.
\newblock \href {https://doi.org/10.18653/v1/D18-1455} {Open domain question
  answering using early fusion of knowledge bases and text}.
\newblock In \emph{Proceedings of the 2018 Conference on Empirical Methods in
  Natural Language Processing}, pages 4231--4242, Brussels, Belgium.
  Association for Computational Linguistics.

\bibitem[{Talmor et~al.(2019)Talmor, Herzig, Lourie, and
  Berant}]{talmor-etal-2019-commonsenseqa}
Alon Talmor, Jonathan Herzig, Nicholas Lourie, and Jonathan Berant. 2019.
\newblock \href {https://doi.org/10.18653/v1/N19-1421} {{C}ommonsense{QA}: A
  question answering challenge targeting commonsense knowledge}.
\newblock In \emph{Proceedings of the 2019 Conference of the North {A}merican
  Chapter of the Association for Computational Linguistics: Human Language
  Technologies, Volume 1 (Long and Short Papers)}, pages 4149--4158,
  Minneapolis, Minnesota. Association for Computational Linguistics.

\bibitem[{Vaswani et~al.(2017)Vaswani, Shazeer, Parmar, Uszkoreit, Jones,
  Gomez, Kaiser, and Polosukhin}]{NIPS2017_3f5ee243}
Ashish Vaswani, Noam Shazeer, Niki Parmar, Jakob Uszkoreit, Llion Jones,
  Aidan~N Gomez, \L~ukasz Kaiser, and Illia Polosukhin. 2017.
\newblock \href
  {https://proceedings.neurips.cc/paper/2017/file/3f5ee243547dee91fbd053c1c4a845aa-Paper.pdf}
  {Attention is all you need}.
\newblock In \emph{Advances in Neural Information Processing Systems},
  volume~30. Curran Associates, Inc.

\bibitem[{Vedantam et~al.(2015)Vedantam, Lawrence~Zitnick, and
  Parikh}]{vedantam2015cider}
Ramakrishna Vedantam, C~Lawrence~Zitnick, and Devi Parikh. 2015.
\newblock Cider: Consensus-based image description evaluation.
\newblock In \emph{Proceedings of the IEEE conference on computer vision and
  pattern recognition}, pages 4566--4575.

\bibitem[{Wang et~al.(2021)Wang, Liu, Zhu, Shou, Gong, Xu, and
  Zeng}]{wang-etal-2021-retrieval-enhanced}
Han Wang, Yang Liu, Chenguang Zhu, Linjun Shou, Ming Gong, Yichong Xu, and
  Michael Zeng. 2021.
\newblock \href {https://doi.org/10.18653/v1/2021.findings-acl.269} {Retrieval
  enhanced model for commonsense generation}.
\newblock In \emph{Findings of the Association for Computational Linguistics:
  ACL-IJCNLP 2021}, pages 3056--3062, Online. Association for Computational
  Linguistics.

\bibitem[{Wang et~al.(2020{\natexlab{a}})Wang, Ren, and
  Leskovec}]{wang2020entity}
Hongwei Wang, Hongyu Ren, and Jure Leskovec. 2020{\natexlab{a}}.
\newblock Entity context and relational paths for knowledge graph completion.
\newblock \emph{arXiv preprint arXiv:2002.06757}.

\bibitem[{Wang et~al.(2020{\natexlab{b}})Wang, Peng, Ilievski, Szekely, and
  Ren}]{pgqa}
Peifeng Wang, Nanyun Peng, Filip Ilievski, Pedro~A. Szekely, and Xiang Ren.
  2020{\natexlab{b}}.
\newblock \href {https://doi.org/10.18653/v1/2020.findings-emnlp.369}
  {Connecting the dots: {A} knowledgeable path generator for commonsense
  question answering}.
\newblock In \emph{Proceedings of the 2020 Conference on Empirical Methods in
  Natural Language Processing: Findings, {EMNLP} 2020, Online Event, 16-20
  November 2020}, volume {EMNLP} 2020 of \emph{Findings of {ACL}}, pages
  4129--4140. Association for Computational Linguistics.

\bibitem[{Wang et~al.(2019)Wang, Kapanipathi, Musa, Yu, Talamadupula,
  Abdelaziz, Chang, Fokoue, Makni, Mattei, and Witbrock}]{wang2019improving}
Xiaoyan Wang, Pavan Kapanipathi, Ryan Musa, Mo~Yu, Kartik Talamadupula, Ibrahim
  Abdelaziz, Maria Chang, Achille Fokoue, Bassem Makni, Nicholas Mattei, and
  Michael Witbrock. 2019.
\newblock \href {https://doi.org/10.1609/aaai.v33i01.33017208} {Improving
  natural language inference using external knowledge in the science questions
  domain}.
\newblock \emph{Proceedings of the AAAI Conference on Artificial Intelligence},
  33(01):7208--7215.

\bibitem[{Xie and Pu(2021)}]{xie2021commonsense}
Yubo Xie and Pearl Pu. 2021.
\newblock How commonsense knowledge helps with natural language tasks: A survey
  of recent resources and methodologies.
\newblock \emph{arXiv preprint arXiv:2108.04674}.

\bibitem[{Yan et~al.(2020)Yan, Raman, Chan, Zhang, Rossi, Zhao, Kim, Lipka, and
  Ren}]{yan2020learning}
Jun Yan, Mrigank Raman, Aaron Chan, Tianyu Zhang, Ryan Rossi, Handong Zhao,
  Sungchul Kim, Nedim Lipka, and Xiang Ren. 2020.
\newblock Learning contextualized knowledge structures for commonsense
  reasoning.
\newblock \emph{arXiv preprint arXiv:2010.12873}.

\bibitem[{Yasunaga et~al.(2021)Yasunaga, Ren, Bosselut, Liang, and
  Leskovec}]{Yasunaga2021QAGNN}
Michihiro Yasunaga, Hongyu Ren, Antoine Bosselut, Percy Liang, and Jure
  Leskovec. 2021.
\newblock \href {https://doi.org/10.18653/v1/2021.naacl-main.45} {{QA}-{GNN}:
  Reasoning with language models and knowledge graphs for question answering}.
\newblock In \emph{Proceedings of the 2021 Conference of the North American
  Chapter of the Association for Computational Linguistics: Human Language
  Technologies}, pages 535--546, Online. Association for Computational
  Linguistics.

\bibitem[{Zellers et~al.(2018)Zellers, Bisk, Schwartz, and
  Choi}]{zellers-etal-2018-swag}
Rowan Zellers, Yonatan Bisk, Roy Schwartz, and Yejin Choi. 2018.
\newblock \href {https://doi.org/10.18653/v1/D18-1009} {{SWAG}: A large-scale
  adversarial dataset for grounded commonsense inference}.
\newblock In \emph{Proceedings of the 2018 Conference on Empirical Methods in
  Natural Language Processing}, pages 93--104, Brussels, Belgium. Association
  for Computational Linguistics.

\bibitem[{Zhang et~al.(2020)Zhang, Liu, Pan, Song, and Leung}]{zhang2020aser}
Hongming Zhang, Xin Liu, Haojie Pan, Yangqiu Song, and Cane Wing-Ki Leung.
  2020.
\newblock Aser: A large-scale eventuality knowledge graph.
\newblock In \emph{Proceedings of The Web Conference 2020}, pages 201--211.

\bibitem[{Zhou et~al.(2020)Zhou, Khanna, Lee, Lin, Ho, Pujara, and
  Ren}]{zhou2020rica}
Pei Zhou, Rahul Khanna, Seyeon Lee, Bill~Yuchen Lin, Daniel Ho, Jay Pujara, and
  Xiang Ren. 2020.
\newblock Rica: Evaluating robust inference capabilities based on commonsense
  axioms.
\newblock \emph{arXiv preprint arXiv:2005.00782}.

\end{thebibliography}
\bibliographystyle{acl_natbib}

\appendix
% \section{Appendix}
\section{Qualitative Comparison}
\label{ap:qcsqa}

Table \ref{table:csqa_qual_pgqa} shows qualitative comparison between \pgsname\ and baselines on the CSQA dataset.

\setul{0.5ex}{0.5ex}
\definecolor{Yellow}{rgb}{1,0.8,0}
\setulcolor{Yellow}

% We conduct a \textbf{human study} wherein we presented evaluators with questions from CSQA dataset with corresponding commonsense paths generated by \pgsname\ and PGQA in an anonymized manner. We asked them to compare the paths based on their contextual relevance with the complete sentence and classify them into one of three categories - 1) ‘\pgsname\ is better than PGQA’, 2) ‘PGQA is better than \pgsname’, 3) ‘Both are of the similar quality’. A total of 150 questions samples were randomly sampled from the test set and presented to 6 evaluators (25 samples each). Following are our observations:

% Number of samples where \pgsname\ is better: 62 (41.33\% of 150 samples) \\
% Number of samples where PGQA is better: 38 (25.33\% of 150 samples) \\
% Number of samples where both are of similar quality: 50 (33.33\% of 150 samples)
 
% This shows that commonsense generated by \pgsname\ is found to be more relevant in human evaluation. Also, if we exclude neutral samples and consider the 100 samples where the path generated by one of either approach is found to be better, \pgsname’s paths are found to be more relevant in 62 samples (62\% of 100 samples) while PGQA’s paths are more relevant in 38 samples (38\% of 100 samples).

\section{Comparison with GPT-2 as backbone language model}
\label{app:gpt_experiments}

We decided to use T5-base as a design choice as we were required to train a text-to-text model where given a sentence as input, the model has to generate the relevant path as output. Since T5-base is a text-to-text generation language model, we felt that it is a suitable choice.

\begin{table}[h]
\centering
\resizebox{0.49\textwidth}{!}{
\begin{tabular}{lcc}
    \toprule
    \multicolumn{1}{c}{}  & \multicolumn{2}{c}{100\% Train} \\
    \multicolumn{1}{c}{\textbf{Methods}}  & \textbf{IHdev} (\%) & \textbf{IHtest} (\%) \\
    \midrule  
    RoBERTa-large (w/o KG) & 73.07~($\pm$0.45) & 68.69 ($\pm$0.56) \\
    \midrule
    + {PGQA w/ GPT-2}  & \underline{77.53}~($\pm$0.47)   & 71.19~($\pm$0.49)  \\
    + {\pgsname\ w/ GPT-2}  & \underline{77.90}~($\pm$0.37)   & 72.67~($\pm$0.18)  \\
    \midrule
    + {PGQA w/ T5-base}  & \underline{77.56}~($\pm$0.32)   & 71.31~($\pm$0.44)  \\
    + {\pgsname\ w/ T5-base}  & \underline{78.15}~($\pm$0.23)   & 72.87~($\pm$0.31)  \\
    \bottomrule 
\end{tabular}
}
% \vspace{-2mm}
\caption{Performance comparison between using T5-base and GPT-2 as backbone language model for PGQA and \pgsname\ for multi-choice QA task on CSQA dataset. 
}
% \vspace{-2mm}
\label{tab:gpt_exps}
\end{table}

\begin{table*}[t]
\resizebox{\textwidth}{!}{%
\renewcommand{\arraystretch}{1.25}%
\begin{tabular}{l|c|c|c|ll}
\hline
\multicolumn{1}{c|}{\textbf{Question}} & \multicolumn{3}{c|}{\textbf{Predictions}} & \multicolumn{2}{c}{\textbf{Generated Paths}}                            \\ \hline
                                       & \textbf{PGQA}     & \textbf{QA-GNN}   & \textbf{Ours}       & \multicolumn{1}{c|}{\textbf{PGQA}} & \multicolumn{1}{c}{\textbf{\pgsname}} \\ \cline{2-6} 
\begin{tabular}[c]{@{}l@{}}Where could you see an advert-\\ -isement while reading news?\end{tabular} & {\color[HTML]{CB0000} television} &
  {\color[HTML]{009901} \begin{tabular}[c]{@{}c@{}}web \\ page\end{tabular}} &
  {\color[HTML]{009901} \begin{tabular}[c]{@{}c@{}}web \\ page\end{tabular}} &
  \multicolumn{1}{l|}{\begin{tabular}[c]{@{}l@{}}(read\_news \textit{hassubevent} read \\ \textit{relatedto} news \textit{atlocation} television) \\ \\ (read\_news \textit{hassubevent} read \\ \textit{relatedto} page)\end{tabular}} &
  \begin{tabular}[c]{@{}l@{}}(spread\_information \textit{\_capableof} \ul{advertisement} \\ \ul{\textit{atlocation} web\_page \textit{usedfor} reading\_news})\\ \\ (news\_article \textit{isa} article \textit{atlocation} web\_page \\ \textit{\_receivesaction} advertisement)\end{tabular} \\ \hline
\begin{tabular}[c]{@{}l@{}}What can years of playing \\ tennis lead to?\end{tabular} &
  {\color[HTML]{CB0000} \begin{tabular}[c]{@{}c@{}}becoming \\ tired\end{tabular}} &
  {\color[HTML]{CB0000} \begin{tabular}[c]{@{}c@{}}becoming \\ tired\end{tabular}} &
  {\color[HTML]{009901} \begin{tabular}[c]{@{}c@{}}tennis \\ elbow\end{tabular}} &
  \multicolumn{1}{l|}{\begin{tabular}[c]{@{}l@{}}(playing\_tennis \textit{causes} \\ becoming\_tired)\\ \\ (play \textit{antonym} fun \textit{usedfor} \\ \ul{playing\_tennis \textit{causes} tennis\_elbow})\end{tabular}} &
  \begin{tabular}[c]{@{}l@{}}(injury \textit{\_hassubevent} playing\_tennis \textit{hasprerequisite} \\ practice\_taking\_care\_of\_sports\_equipment)\\ \\ (\ul{playing\_tennis \textit{hassubevent} injury \textit{hasprerequisite}} \\ \ul{practice} \textit{\_hasfirstsubevent} be\_better\_at\_new\_things)\end{tabular} \\ \hline
\begin{tabular}[c]{@{}l@{}}A person writes a check to a clerk, \\ where does the clerk put them?\end{tabular} &
  {\color[HTML]{CB0000} \begin{tabular}[c]{@{}c@{}}desk \\ drawer\end{tabular}} &
  {\color[HTML]{009901} \begin{tabular}[c]{@{}c@{}}cash \\ register\end{tabular}} &
  {\color[HTML]{009901} \begin{tabular}[c]{@{}c@{}}cash \\ register\end{tabular}} &
  \multicolumn{1}{l|}{\begin{tabular}[c]{@{}l@{}}(put \textit{relatedto} desk \textit{partof} drawer)\\ \\ (check \textit{relatedto} cash \textit{relatedto} register)\\ \\ (write \textit{relatedto} desk \textit{partof} drawer)\end{tabular}} &
  \begin{tabular}[c]{@{}l@{}}(make\_payments \textit{\_capableof} \ul{clerk desires check} \\ \ul{\textit{\_atlocation} cash\_registers} \textit{\_usedfor} to\_pay\_for\_goods)\\ \\ (cash\_registers \textit{\_usedfor} clerk \textit{isa} person \textit{desires} \\ clean\_house \textit{hasprerequisite} put\_things\_into\_places)\end{tabular} \\ \hline
\begin{tabular}[c]{@{}l@{}}Where could you find some large \\ pieces of paper that are not for sale?\end{tabular} &
  {\color[HTML]{CB0000} \begin{tabular}[c]{@{}c@{}}office \\ supply \\ store\end{tabular}} &
  {\color[HTML]{CB0000} \begin{tabular}[c]{@{}c@{}}cabinet \end{tabular}} &
  {\color[HTML]{009901} \begin{tabular}[c]{@{}c@{}}artist's \\ studio\end{tabular}} &
  \multicolumn{1}{l|}{\begin{tabular}[c]{@{}l@{}}(large \textit{relatedto} note \textit{relatedto} \\ paper \textit{relatedto} office\_supply)\\ \\ (pieces \textit{relatedto} part \textit{relatedto} \\ paper \textit{relatedto} office\_supply)\end{tabular}} &
  \begin{tabular}[c]{@{}l@{}}(shredded\_paper \textit{usedfor} sale \textit{\_hassubevent} \\ \ul{buying\_products \textit{\_nothasproperty} artist\_studio})\\ \\ (write\_letters \textit{\_usedfor} paper \textit{receivesaction} \\ sell\_for\_money \textit{atlocation} store)\end{tabular} \\ \hline
\begin{tabular}[c]{@{}l@{}}What do humans take in while \\ breathing?\end{tabular} &
  {\color[HTML]{CB0000} air} &
  {\color[HTML]{009901} oxygen} &
  {\color[HTML]{009901} oxygen} &
  \multicolumn{1}{l|}{\begin{tabular}[c]{@{}l@{}}(humans \textit{relatedto} air)\\ \\ (breathing \textit{hassubevent} air)\\ \\ (human \textit{relatedto} \ul{breathing} \\ \ul{\textit{hassubevent} oxygen})\end{tabular}} &
  \begin{tabular}[c]{@{}l@{}}(breathing \textit{hassubevent} inhale \textit{motivatedbygoal} \\ \ul{fresh\_air \textit{\_atlocation} oxygen})\\ \\ (inhaling \textit{\_hassubevent} breathing \textit{causes} life \\ \textit{\_usedfor} living\_life \textit{hasprerequisite} good\_health)\end{tabular} \\ \hline
\end{tabular}%
}
% \caption{Examples of commonsense inferences obtained for different input forms of the same question from \pgsname\ when trained with different values of $p_{mask}$. Potential answers which are observed in a path are \colorbox{yellow}{highlighted}, while context-enriching concepts are \textcolor{cyan}{\textbf{coloured}}.}
\caption{Comparison between predictions made by PGQA~\citep{pgqa}, QA-GNN~\citep{Yasunaga2021QAGNN}, and \pgsname\ on a subset of CSQA's in-house test set~\citep{talmor-etal-2019-commonsenseqa}. Commonsense paths that are responsible for the corresponding predictions are also given for both the path-based models. \ul{Underlined} portions represent the meaningful path sub-structures which direct the overlying model towards the correct answer.}
\label{table:csqa_qual_pgqa}
\end{table*}

To empirically establish that improvements over PGQA are not due to using T5-base instead of GPT-2, we performed an experiment to replace T5-base with GPT-2 as the backbone language model of CoSe-Co. We train GPT-2 using the same sentence-path dataset as we used for T5-base by providing it as input the sentence followed by a [SEP] token and adapting GPT-2 to generate the corresponding path. Additionally, we also experiment with replacing the language model in PGQA from GPT-2 to T5-base. Table \ref{tab:gpt_exps} summarises the results obtained for multi-choice QA on CSQA where it can be seen that using GPT-2 vs T5 does not lead to noticeable changes in the performance. The test accuracy attained by CoSe-Co with T5-base is 72.87\% which is almost the same as for CoSe-Co with GPT-2: 72.67\%. A similar observation is seen for PGQA where using T5-base backbone gives 71.31\% and using GPT-2 gives 71.19\%. Further, we would like to highlight that CoSe-Co with GPT-2 backbone attains 72.67\% accuracy and performs better than PGQA with GPT-2 (71.19\%).

Based on these observations, we can conclude that performance gains of CoSe-Co over PGQA are not due to using different backbone but because CoSe-Co is trained over semantically related sentence-commonsense pairs that enables it to generate contextually more relevant commonsense.

\section{Entity masking while training \pgsname}
\label{ap:mask}
Table \ref{table:path_mask_eval} shows the various kinds of paths obtained from \pgsname\ when trained with different values of $p_{mask}$, across the same original question. A number of observations can be made. First, the paths obtained from the variant which is trained without any masking ($p_{mask} = 0.0$) produces inferences that enrich the overall context of certain entities in question but do not necessarily capture the inter-relation between them and thus the overall intention of the question. With the configurations that are trained with $p_{mask} \neq 0$, the various paths capture the overall context in an answer-oriented manner. These configurations also allow us to mask concepts in the original question such that \pgsname\ can exploit the unmasked entities to direct its generated paths in a manner that best suit the blank. This is evident from the second half of Table \ref{table:path_mask_eval}. When the interrogative element is masked in the first example, the paths are directed towards actually finding the best answer, while when `Google maps' is replaced in the third example, the paths are clearly focused on predicting concepts related to GPS systems. 

% \colorbox{yellow}{
% \textcolor{ProcessBlue}{\textbf{
\begin{table*}[h]
\centering
\resizebox{\textwidth}{!}{
\begin{tabular}{@{}lll@{}}
\toprule
\textbf{Input} & \textbf{\makecell{\pgsname\ \\ Variant}} & \textbf{\pgsname\ Outputs} \\ \midrule
\multirow{5}{*}{\begin{tabular}[c]{@{}l@{}}Google maps and other GPS \\services have replaced what?\end{tabular}} & \multirow{5}{*}{$p_{mask}=0.0$} 
 & (orienteering \textit{\_usedfor} map \textit{\_isa} \textcolor{cyan}{\textbf{navigation}} \textit{\_hascontext} GPS \textit{\_product} gsrc) \\
 &  & (satellite\_gear \textit{hascontext} maps \textit{capableof} \textcolor{cyan}{\textbf{help\_find\_places}} \textit{\_usedfor} map\_store\_in\_city) \\
 &  & (similar\_to\_gps \textit{\_hasproperty} google \textit{hascontext} maps \textit{capableof} \textcolor{cyan}{\textbf{show\_live\_places}}) \\
 &  & (geospatial \textit{\_isa} navigation \textit{\_usedfor} google \textit{hascontext} maps \textit{capableof} show\_people\_places) \\
 \midrule
 \multirow{5}{*}{\begin{tabular}[c]{@{}l@{}}Google maps and other GPS \\services have replaced what?\end{tabular}} & \multirow{5}{*}{$p_{mask}=1.0$} 
 & (gps \textit{hascontext} \textcolor{cyan}{\textbf{navigation}} \textit{\_usedfor} satellite \textit{atlocation} orbital\_stations\_in\_space) \\
 &  & (satellite\_navigation \textit{\_usedfor} \colorbox{yellow}{compass} \textit{capableof} \textcolor{cyan}{\textbf{point\_north}} \textit{\_hassubevent} \textcolor{cyan}{\textbf{driving\_car}}) \\
 &  & (navigation \textit{\_usedfor} satellite \textit{\_isa} gps \textit{hascontext} navigation \textit{\_entails} competition) \\
 &  & (\textcolor{cyan}{\textbf{location\_where\_go}}\_gallery \textit{\_definedas} navigation \textit{\_usedfor} satellite \textit{\_hascontext} gps) \\
 \midrule
 \multirow{5}{*}{\begin{tabular}[c]{@{}l@{}}Google maps and other GPS \\services have replaced what?\end{tabular}} & \multirow{5}{*}{$p_{mask}=0.5$} 
 & (\colorbox{yellow}{compass} \textit{\_isa} google\_maps \textit{usedfor} \textcolor{cyan}{\textbf{navigation}} \textit{\_hascontext} gps \textit{isa} navigating\_map) \\
 &  & (\textcolor{cyan}{\textbf{location\_where\_go\_camping}} \textit{\_isa} location \textit{\_usedfor} map \_product google\_maps) \\
 &  & (satellite\_gear \textit{\_isa} GPS \textit{usedfor} navigation \textit{\_hascontext} gps \textit{isa} navigating\_map) \\
 &  & (navigation\_maps \textit{\_usedfor} map \textit{\_isa} navigation\_map \textit{hascontext} navigation) \\
 \midrule
 \multirow{5}{*}{\begin{tabular}[c]{@{}l@{}}Google maps and other GPS \\services have replaced what?\end{tabular}} & \multirow{5}{*}{$p_{mask}=0.33$} & (orienteering \textit{\_usedfor} maps \textit{\_isa} google\_maps \textit{\_hasprerequisite} looking\_through\_telescope) \\
 &  & (\textcolor{cyan}{\textbf{location\_where\_go\_shopping}} \textit{\_definedas} where\_go\_shopping \textit{\_usedfor} map) \\
 &  & (navigation\_maps \textit{\_isa} maps \textit{\_usedfor} satellite \textit{locatednear} planet) \\
 &  & (satellite\_navigation \textit{\_usedfor} maps \textit{\_hascontext} google\_maps \textit{capableof} show\_locations) \\
 \midrule
 \midrule
\multirow{5}{*}{\begin{tabular}[c]{@{}l@{}}Google maps and other GPS \\services have replaced {[}\colorbox{yellow}{MASK}{]}.\end{tabular}} & \multirow{5}{*}{$p_{mask}=0.33$} & (gps \textit{hascontext} maps \textit{\_usedfor} satellite \textit{locatednear} planet) \\
 &  & {(navigation\_maps \textit{isa} navigation \textit{\_usedfor} compass \textit{capableof} \textcolor{cyan}{\textbf{point\_north\_handle}})} \\
 &  & (satellite\_navigation \textit{\_usedfor} \colorbox{yellow}{compass} \textit{capableof} point\_north\_or\_south\_hemispheres) \\
 &  & (location\_where\_go\_if\_near\_beach \textit{\_definedas} map \textit{usedfor} navigation \textit{\_mannerof} sport) \\
 \midrule
\multirow{5}{*}{\begin{tabular}[c]{@{}l@{}}Google maps and other GPS \\services have {[}\colorbox{yellow}{MASK}{]} what?\end{tabular}} & \multirow{5}{*}{$p_{mask}=0.33$} & (orienteering \textit{\_usedfor} map \textit{\_isa} google\_maps \textit{\_hascontext} gps) \\
 &  & {(\textcolor{cyan}{\textbf{location\_where\_go\_if\_need\_to}} \textit{\_definedas} location \textit{\_isa} map \textit{usedfor} \textcolor{cyan}{\textbf{information}})} \\
 &  & (located\_in\_latin\_america \textit{\_receivesaction} israel \textit{\_language} latin\_america) \\
 &  & (navigation\_maps \textit{usedfor} find\_place \textit{\_hasprerequisite} go\_to\_market) \\
 &  & (satellite\_navigation \textit{\_usedfor} maps \textit{capableof} \textcolor{cyan}{\textbf{show\_locations\_and\_routes}}) \\
 \midrule
\multirow{5}{*}{\begin{tabular}[c]{@{}l@{}}{[}\colorbox{yellow}{MASK}{]} and other GPS \\services have replaced what?\end{tabular}} & \multirow{5}{*}{$p_{mask}=0.33$} & (navigation\_system \textit{\_isa} GPS \textit{hascontext} astronomy \textit{\_field} edmond\_halley) \\
 &  & (location\_where\_go\_if\_in\_accident \textit{\_usedfor} map \textit{\_atlocation} GPS\_systems) \\
 &  & {(\textcolor{cyan}{\textbf{radio\_frequency\_messaging}} \textit{\_isa} GPS \textit{hasproperty} useful)} \\
 &  & (receiver \textit{partof} radio \textit{\_isa} gps \textit{hascontext} navigation \textit{\_usedfor} \colorbox{yellow}{compass}) \\
\bottomrule
\end{tabular}
}
\caption{Examples of commonsense inferences obtained for different input forms of the same question from \pgsname\ when trained with different values of $p_{mask}$. Potential answers which are observed in a path are \colorbox{yellow}{highlighted}, while context-enriching concepts are \textcolor{cyan}{\textbf{coloured}}.}
\label{table:path_mask_eval}
\end{table*}

\section{Details of PGQA Baseline}
\label{app:pgqa_details}
PGQA \citep{pgqa} leverages the commonsense paths generated by their path generator module along with the question and candidate answer choices to perform multi-choice QA on CSQA dataset \citep{talmor-etal-2019-commonsenseqa}. Specifically, given a question $q$ with corresponding candidate answer choices set $C = \{c_1, \dots , c_n\}$, the PGQA framework generates commonsense inferences for each pair of answer choice $c_i$ and entities extracted from $q$. A total of $k$ paths corresponding to each answer choice $c_i$ are obtained to get a resultant set of paths - $P_{q-c_i}$. Further, an average over the hidden representations corresponding to sequence of decoded tokens from the final layer of their path generator decoder are used as path embedding and combined as - $H_{S} \in R^{k \times h_{D}}$ to represent the paths in $P_{q-c_i}$. Following this, they augment the choice into $q$ by replacing the interrogative phrase in $q$ with $c_i$ to obtain $q'$. For instance, given the question `Google maps and other GPS services have replaced what?', the answer choice `atlas' is augmented into the question as: `Google maps and other GPS services have replaced \textit{atlas}.'

To embed the augmented question and corresponding answer choice, they use a pre-trained LM encoder $E$ (such as RoBERTa \citep{roberta}) to embed the query - `[CLS] $q'$ [SEP] $c_i$' corresponding to $c_i$. The representation corresponding to [CLS] token is extracted from the final hidden layer as $h_{US} \in R^{h_{E}}$. In order to leverage relevant knowledge from the generated commonsense inferences, the question and choice embeddings are used to attend over generated paths as:

\begin{equation*}
    \alpha_p = Softmax(tanh(H_{S}W^{A}) h_{US}) \notag\\
\end{equation*}

\begin{equation*}
h_{S'} = \sum\limits_{h \in H_{S}} \alpha_{p}^{h} \cdot h    
\end{equation*}

% $$    
% \end{center}

% and aggregate over them to obtain the final structured embeddings
where, $W^A \in R^{h_{D} \times h_E}$, $\alpha_p \in R^{k}$ and $h_{S'} \in R^{h_{D}}$. Finally, a linear layer is applied over the concatenation of $\{h_{US}, h_{S'}\}$ to project it as a scalar. A softmax is taken over concatenation of scalars obtained corresponding to each answer choice to obtain their likelihood followed by cross entropy loss for training.

\section{Further Implementation Details}
\label{ap:imp_det}

\subsection{Relation Heuristics}
\label{ap:rel_heur}
As mentioned in \cref{sec:s_to_p}, we employ heuristics on the basis of contained relations to perform filtering of ConceptNet paths. Particularly, we use the following rules: 
\begin{enumerate}
    \item We discard any path that uses the same two relations to connect any three neighbouring entities occurring in it. That is, for any sub-path $\{e_i, r_i, e_{i+1}, r_{i+1}, e_{i+2}\}$ in a given path $p$, we only consider $p$ as a part of our dataset if $r_i \neq r_{i+1}$. 
    \item Following \citep{pgqa}, we do not consider paths that contain any relations from the set $\{$\textit{HasContext}, \textit{RelatedTo}, \textit{Synonym}, \textit{Antonym}, \textit{DerivedFrom}, \textit{FormOf}, \textit{EtymologicallyDerivedFrom}, \textit{EtymologicallyRelatedTo}$\}$. We observed that entities connected through these relations were often largely dissimilar and thus not useful for our case.
\end{enumerate}

% \subsection{KG Completion}
% As discussed in \cref{sec:analyse_p}, we use \pgsname\ for the task of KG completion where given a test triple `h, r and t', we give h and r as input to \pgsname\ and also condition the decoder with input `h r' and then generate the next entity. To compute the accuracy, we perform matching between generated entity and ground truth tail entity in the triple. To perform comparison with COMET~\citep{comet} we take their code and pre-trained model from \href{https://github.com/atcbosselut/comet-commonsense}{here}.

\subsection{Multi-Choice QA}
In \cref{sec:down_csqa}, we discuss commonsense question answering task where we use framework developed by \citet{pgqa} and just replace the commonsense knowledge used by them with the paths generated by \pgsname. We use the same hyper-parameters as used by them and mention them here for reference. The model is trained on a batch size of 16, dropout of 0.1 for 15 epochs. A learning rate of 2e-6 is used for encoder LM (Roberta-large) used for embedding question and choice context and an lr of 1e-3 is used for remaining path attention and classification layer parameters. We perform the evaluation on CSQA~\citep{talmor-etal-2019-commonsenseqa} dataset downloaded from \href{https://www.tau-nlp.org/commonsenseqa}{here}. The train split comprises of 8,500, dev split contains 1,221 and in-house test split contains 1,241 samples.

\subsection{OpenCSR}
In this section, we discuss the implementation details used for OpenCSR in \cref{sec:down_opencsr}. The dataset has been downloaded from \href{https://github.com/yuchenlin/OpenCSR/tree/main/drfact_data}{here}. The training splits of ARC, QASC, and OBQA datasets comprises of $5355$, $6883$, and $4199$ samples respectively while the development split comprises of $562$, $731$, and $463$ samples respectively. The test set is hidden and authors who proposed the task with reformulated dataset are yet to set up a leaderboard on the hidden test set. They run their proposed model DrFact (which is based on BERT-base and is the current state-of-the-art on this task) on a single seed which takes about $\sim$2-3 days to train one model on a given dataset. While fine-tuning T5-base (with and without \pgsname\ knowledge), we train the model for 5 epochs with a learning rate of 5e-4, weight decay of 0.01 and batch size 8 using AdamW optimizer~\citep{adamw2017}.

\subsection{Paraphrase Generation}
\label{sec:other_tasks_app}
For paraphrase generation on MRPC~\citep{dolan-brockett-2005-automatically} dataset, we fine-tune T5-base (with and without \pgsname\ knowledge) at a learning rate of 5e-4 for 5 epochs with weight decay of 0.01 and 4 gradient accumulation steps using AdamW~\citep{adamw2017} optimizer. The training set of MRPC comprises of 2,661 paraphrases while the test set comprises of 1,088 paraphrases. The dataset has been downloaded from \href{https://www.microsoft.com/en-us/download/details.aspx?id=52398}{here}.

\section{Further Analysis of Generated Paths}
\label{app:disc_gen_paths}
\begin{itemize}
    \item \textbf{Correctness of Novel Triples} : Since there is no ground truth to check the correctness of triple comprising of novel entities, we attempt to evaluate them by leveraging a commonsense knowledge base completion model - Bilinear AVG~\citep{li-etal-2016-commonsense} which has been shown to achieve an accuracy of 92.5\% on knowledge completion task and is also used to score triples. We extract triples comprising of at least one novel entity from the paths generated by \pgsname\ for the test split of sentence-path dataset and provide the triple to Bilinear AVG to obtain a score. The average score over all the triples is 0.414 (on a scale of 0 to 1).
    
    \item Further, we perform KG completion (predicting tail entity given head entity and relation of a KG triple) using \pgsname\ since it generates paths which essentially comprise of triples. We compare the performance with COMET~\citep{comet}. We consider test split of sentence-path dataset comprising of $11,264$ paths and extract triples. We filter out triples appearing in training paths of \pgsname\ and train set triples of COMET yielding $717$ test triples in total. \pgsname\ achieves an accuracy of $24.12\%$ which is significantly better than COMET which provides accuracy of $9.76\%$. To perform comparison with COMET~\citep{comet} we take their code and pre-trained model from \href{https://github.com/atcbosselut/comet-commonsense}{here}.
    
    \item In Figure \ref{fig:pg_eval}(b), greedy decoding cannot be compared for diversity with other methods since it generates only a single unique path.
    
    \item Since generated paths diversity estimates can be affected by path length, we measure the standard deviation of the number of entities in paths generated corresponding to test split sentences and found it to be 0.76 which shows that variance in the lengths of generated paths is very low (<1) and hence, the diversity of 0.43 (on a scale 0 to 1) attained by \pgsname\ is not due to length bias.
\end{itemize}

\section{Ethics statement}
\label{ap:ethics}
% \item In our work, we have mainly focused on developing a model that generates contextually more relevant knowledge which can be generically used in different NLP tasks in a scalable manner. However, we did not focus on exploring how the generated knowledge can be utilised better in downstream tasks to achieve even better gains in addition to common augmentation techniques discussed in the paper.

\begin{itemize}
    \item The sentence - commonsense dataset created to train CoSe-Co has been derived using standardized Wikipedia Corpus and ConceptNet knowledge graph which are publicly available and commonly used without containing any info/text that could potentially lead to risk impacts.
    \item We have used open source Wikipedia corpus and ConceptNet which are publicly available and already standardized for research works.
    \item The links to all the previous works, their provided open-source github repos, artifacts and datasets have been provided in appropriate sections where they are discussed/used/compared along with their citations (Sections - \ref{sec:rel_work}, \ref{sec:down}, Appendix \ref{ap:imp_det} etc.). The links to any resources used provide permissions to use them for our research work.
\end{itemize}

\end{document}